%% file: neurips_data_2022.tex
\documentclass{article}

    \PassOptionsToPackage{numbers, compress}{natbib}

    \usepackage[final]{neurips_data_2022}

\usepackage[utf8]{inputenc} 
\usepackage[T1]{fontenc}    
\usepackage{hyperref}       
\usepackage{url}            
\usepackage{booktabs}       
\usepackage{amsfonts}       
\usepackage{nicefrac}       
\usepackage{microtype}      
\usepackage{xcolor}         

\usepackage{times}
\usepackage{latexsym}
\usepackage{graphicx}
\usepackage{adjustbox}
\usepackage{multirow, multicol}
\usepackage{caption}
\usepackage{multicol}
\usepackage{makecell}
\usepackage{amsmath}
\usepackage{tabularray}

\usepackage{arydshln}
\usepackage{enumitem}
\usepackage{pifont}
\newcommand{\cmark}{\ding{51}}
\newcommand{\xmark}{\ding{55}}

\title{EHRSQL: A Practical Text-to-SQL Benchmark for Electronic Health Records}

\author{
    Gyubok Lee$\textsuperscript{1}$, 
    Hyeonji Hwang$\textsuperscript{1}$, 
    Seongsu Bae$\textsuperscript{1}$, 
    Yeonsu Kwon$\textsuperscript{2}$, \\
    \textbf{Woncheol Shin$\textsuperscript{1}$}, 
    \textbf{Seongjun Yang$\textsuperscript{1}$},
    \textbf{Minjoon Seo$\textsuperscript{1}$},  
    \textbf{Jongyeup Kim$\dagger$}, 
    \textbf{Edward Choi$\textsuperscript{1}$} \\
    KAIST, Daejeon$\textsuperscript{1}$ 
    Seoul Women’s University, Seoul$\textsuperscript{2}$ \\
    College of Medicine, Konyang University, Daejeon$\dagger$\\
    \texttt{\{gyubok.lee,localh,seongsu,swc1905,seongjunyang,} \\
    \texttt{minjoon,edwardchoi\}@kaist.ac.kr}$\textsuperscript{1}$ \\
    \texttt{dustn1259@swu.ac.kr}$\textsuperscript{2}$, 
    \texttt{jykim@kyuh.ac.kr}$\dagger$
}

\begin{document}

\maketitle

\begin{abstract}
We present a new text-to-SQL dataset for electronic health records (EHRs). The utterances were collected from 222 hospital staff members, including physicians, nurses, and insurance review and health records teams. To construct the QA dataset on structured EHR data, we conducted a poll at a university hospital and used the responses to create seed questions. We then manually linked these questions to two open-source EHR databases, MIMIC-III and eICU, and included various time expressions and held-out unanswerable questions in the dataset, which were also collected from the poll. Our dataset poses a unique set of challenges: the model needs to 1) generate SQL queries that reflect a wide range of needs in the hospital, including simple retrieval and complex operations such as calculating survival rate, 2) understand various time expressions to answer time-sensitive questions in healthcare, and 3) distinguish whether a given question is answerable or unanswerable. We believe our dataset, EHRSQL, can serve as a practical benchmark for developing and assessing QA models on structured EHR data and take a step further towards bridging the gap between text-to-SQL research and its real-life deployment in healthcare. EHRSQL is available at \url{https://github.com/glee4810/EHRSQL}.
\end{abstract}

\section{Introduction}
\label{sec:introduction}

Electronic health records (EHRs) are relational databases that store a patient's entire medical history in the hospital. From hospital admission to patient treatment and discharge, all medical events that happened in the hospital are recorded and stored in the EHR. The records cover a wide range of clinical knowledge, from individual-level information to group-level insight, in various forms, including tables, text, and images~\cite{johnson2016mimic, johnson2019mimic, goldberger2000physiobank, johnson2021mimic}. As a vast and comprehensive knowledge base, hospital staff, including physicians, nurses, and administrators, constantly interact with EHRs to store and retrieve patient information to make better clinical decisions~\cite{wang2020text, bardhan2022drugehrqa}.

Meanwhile, most hospital staff interact with the EHR database using pre-defined rule conversion systems. To look for information beyond the rules, one needs to undergo special training to modify and extend the system~\cite{wang2020text}. As a result, it is a massive bottleneck for the users to fully utilize the information stored in the EHR. An alternative way to tackle this problem is to build a system that can automatically translate questions directly into the corresponding SQL queries. Then, the users can simply type or verbally ask their questions to the system, and it will return the answers without going through any complicated process. Such systems will unleash the potential of EHR data and tremendously speed up any task involving data retrieval from the database.

Existing datasets that tackle question answering (QA) over structured EHR data are MIMICSQL~\cite{wang2020text} and emrKBQA~\cite{raghavan2021emrkbqa}. MIMICSQL is the first dataset for healthcare QA on MIMIC-III~\cite{johnson2016mimic}, where the questions are automatically generated with pre-defined templates. emrKBQA is another dataset on MIMIC-III derived from emrQA~\cite{pampari2018emrqa}, a clinical reading comprehension dataset. However, through a poll at a university hospital, we discover that the existing datasets are far from fulfilling the actual needs in the hospital workplace in several aspects.

We present EHRSQL\footnote{The dataset is distributed under the CC BY 4.0 license.}, a new large-scale text-to-SQL dataset linked to two open-source EHR databases—MIMIC-III~\cite{johnson2016mimic} and eICU~\cite{pollard2018eicu}. To the best of our knowledge, our work is the first EHR QA dataset that reflects the diverse needs of hospital staff while introducing practical issues in deploying healthcare QA systems. The unique challenges posed by our dataset are threefold:

\begin{itemize}[leftmargin=5.5mm]
	\item \textbf{Wide range of questions:} We conducted a poll at a university hospital to collect questions frequently asked on structured EHR data. The total number of respondents was 222 people with varying years of experience in their professions (see Figure~\ref{figure:data_source}). After filtering and templatizing the responses, the resulting templates cover a wide range of questions, including retrieving patient records (\textit{e.g.}, vital signs measured and hospital cost) and conducting complex group-level operations (\textit{e.g.}, retrieving the top \texttt{N} medications prescribed after being diagnosed with a disease and calculating the \texttt{N}-year survival rate). See Table~\ref{tab:sample-table} for more examples.
	\item \textbf{Time-sensitive questions:} Based on the poll, we learned that real-world questions in the hospital workplace are rich in time expressions, as time is one of the most crucial aspects of healthcare. To reflect this in the dataset, we systematically categorized time into multiple expression types (\textit{e.g.}, absolute, relative, and mixed), units (\textit{e.g.}, hospital visit, month, and day), and interval types (\textit{e.g.}, since, until, and in). Then, we combined the categorized time with the question templates to simulate time-sensitive questions asked in the hospital. More details are discussed in Section~\ref{ssec:time_template}.
	\item \textbf{Trustworthy QA systems:} Developing trustworthy systems is crucial for the adoption of AI in healthcare. Likewise, QA systems need to return only accurate answers and refuse to answer questions beyond their capabilities. To test this, we include unanswerable questions in the dataset, utilizing the remaining utterances from the poll result. These utterances are unanswerable due to incompatibility with the database schema or requiring external domain knowledge.
\end{itemize}

\section{Related Works}

\paragraph{EHR QA}
\citet{pampari2018emrqa} proposed emrQA for question answering on clinical notes based on physicians' frequently asked questions. Later, \citet{raghavan2021emrkbqa} proposed emrKBQA\footnote{emrKBQA is yet to be publicly released at the time of this work.}, which adapted emrQA to structured patient records in MIMIC-III. Due to the nature of the original dataset targeting outpatients, the scope of the questions in emrKBQA is limited to mostly asking about patients' test results~\cite{raghavan2021emrkbqa}. 
Recently, \citet{wang2020text} proposed MIMICSQL, which first tackles EHR QA with the text-to-SQL generation task. To construct the dataset, they automatically generated questions based on pre-defined rules and filtered them through crowd-sourcing.
However, MIMICSQL contains a limited scope of questions and simple SQL queries restricted to five tables, which are far from the actual use case in the hospital~\cite{park2021knowledge}. EHRSQL, on the other hand, originates from the actual poll result and covers a variety of real-world questions (spanning 13.5 tables on average) frequently asked on structured EHR data (see Table~\ref{tab:sample-table}).

\begin{table}[t!]
\caption{Sample utterances in EHRSQL (SQL queries are labeled differently according to the MIMIC-III and eICU schemas).}
\label{tab:sample-table}
\centering
\renewcommand{\arraystretch}{1.2}
\begin{adjustbox}{width=\columnwidth,center}  
\begin{tabular}{ccllc}
\toprule
\multicolumn{2}{c}{\textbf{Type}}     & \multicolumn{1}{c}{\textbf{Sample question}} & \textbf{Department} \\
\midrule
\multirow{10}{*}{\makecell{ Question \\ diversity } } & Demographics & Tell me the \textbf{birthdate} of patient 92721?  & Nursing \\
\cmidrule(r){2-4}
& Prescription & When was the first \textbf{prescription} time of patient 20000? & Nursing \\    
\cmidrule(r){2-4}
& \multirow{2}{*}{Vital sign} & \multirow{2}{*}{What was the last \textbf{arterial bp [systolic]} for patient 23042?} & Physician, Nursing, \\  
& & & Health records, Other \\
\cmidrule(r){2-4}
& Cost & What is the \textbf{average total hospital cost} that involves non-invasive mech vent? & Insurance review\\    
\cmidrule(r){2-4}
& Survival rate & What are the five diagnoses which have the lowest \textbf{four year survival rate}? & Nursing \\         
\cmidrule(r){2-4}
& Longitudinal & What are the five most commonly \textbf{prescribed drugs} for patients who have been  &  \multirow{2}{*}{Physician, Nursing} \\ 
& statistics & \textbf{diagnosed with hypotension nos} earlier since 2104 within 2 months? &  \\
\midrule
\multirow{8}{*}{\makecell{ Time-sensitive \\ question } }  & \multirow{3}{*}{Absolute} & Is the heart rate of patient 2518 measured \textbf{at 2105-12-31 12:00:00} less than &  \multirow{2}{*}{Physician} \\
& &  the value measured \textbf{at 2105-12-31 09:00:00}? \\
\cmidrule(r){3-4}
& & Tell me the last specimen test given to patient 51177 in \textbf{10/2105?}  & Nursing \\
\cmidrule(r){2-4}
& Relative & Tell me patient 6990's daily average gastric meds intake \textbf{on the last ICU visit}. & - \\
\cmidrule(r){2-4}
& \multirow{2}{*}{Mixed} & Tell me the medication that patient 3929 has been prescribed for the first time & \multirow{2}{*}{Nursing} \\
& & \textbf{in 05/this year}. \\
\midrule
\multirow{3.5}{*}{\makecell{ Unanswerable \\ question } } & \multirow{2}{*}{\makecell{ Requiring \\ external knowledge } } & What is the name of the \textbf{medication which should not be administered} during & \multirow{2}{*}{Nursing} \\
& & the contrast arteriogram-leg treatment? \\
\cmidrule(r){2-4}
& Beyond DB schema  &  When is the \textbf{next earliest hospital visit} of patient 73652? & Nursing \\
\bottomrule
\end{tabular}
\end{adjustbox}
\label{tab:sample_data}
\end{table}

\paragraph{Semantic parsing datasets} 

Semantic parsing has been one of the most active areas of research in natural language processing over the past few decades~\cite{hemphill-etal-1990-atis, zelle1996learning, yaghmazadeh2017sqlizer, finegan2018improving}. It involves converting natural language utterances to logical form representations, which are often executable programs such as Python scripts and SQL queries.
WikiSQL~\cite{zhong2017seq2sql} and Spider~\cite{yu2018spider} are leading datasets that assess semantic parsing models on unseen databases. WikiSQL provides a wide range of database domains extracted from HTML tables in Wikipedia, but it only contains simple SQL queries and single tables. Spider aims to tackle complex queries, including joining and set operations, on multiple tables in multiple database domains. Recently, more realistic datasets have been proposed to bridge the gap between academic and practical settings in semantic parsing. KaggleDBQA~\cite{lee2021kaggledbqa} utilizes real-world databases from Kaggle and constructs domain-specific questions without relying on the database schema. SEDE~\cite{hazoom2021text} deals with complex SQL queries that Stack Exchange users ask in real life. Our motivation aligns with theirs in that additional challenges arise in real-world healthcare QA systems.

\paragraph{Unanswerable questions in QA}

In the early days of question answering, unanswerable questions were generated via distant supervision~\cite{joshi-etal-2017-triviaqa, clark2017simple}, rule-based editing~\cite{jia-liang-2017-adversarial}, or adversarial creation by crowd workers~\cite{rajpurkar2018know}. Yet, in semantic parsing, most works assume that all input questions are valid and can be answered; however, this is not true in practice~\cite{zhang2020did}. Retrieving answers to all the input questions is not always desirable for the model to ensure system reliability. To address this, \citet{zhang2020did} utilized other text-to-SQL and chit-chat datasets to construct unanswerable questions and posed the problem as a classification task to detect unanswerable questions. Our work, on the other hand, addresses both system reliability and semantic parsing simultaneously and the unanswerable questions are naturally collected through the poll.
In task-oriented dialog systems, the same task has been tackled in the name of Out-Of-Domain (OOD) detection, and several recent works approach this problem without explicitly showing OOD samples to the model during training for practical reasons. Following their setup, our unanswerable questions are only included in the validation and test sets.

\section{Dataset Construction}
\label{sec:dataset_construction}

\begin{figure}[t!]
    \centering
    \begin{minipage}{0.49\textwidth}
        \centering
        \includegraphics[width=1.0\textwidth]{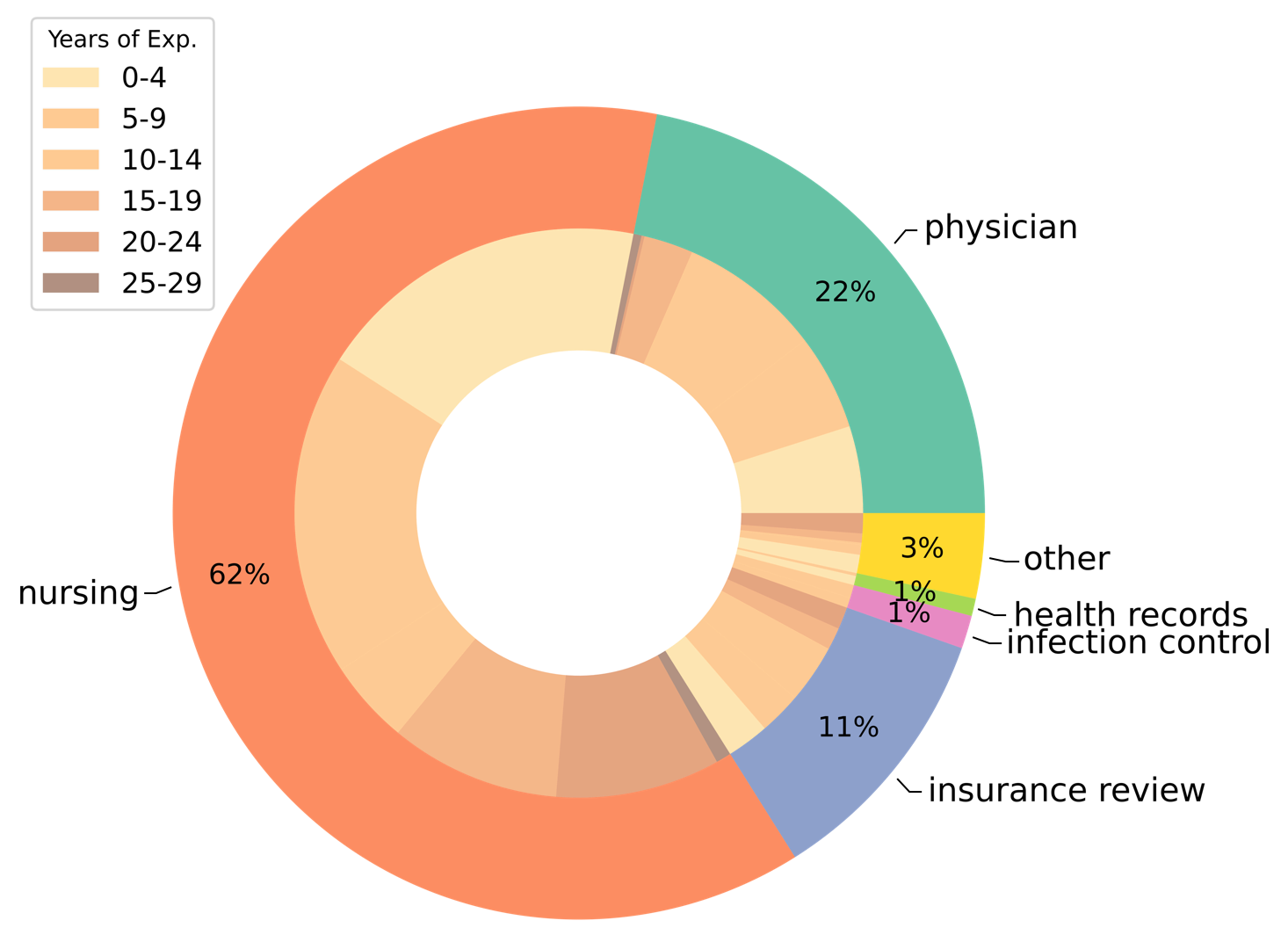}
        \vspace{0.01cm}
        \caption{Demographics of the 222 respondents by hospital departments and the years of experience.}
        \label{figure:data_source}
    \end{minipage}\hfill
    \begin{minipage}{0.49\textwidth}
        \centering
        \includegraphics[width=1.0\textwidth]{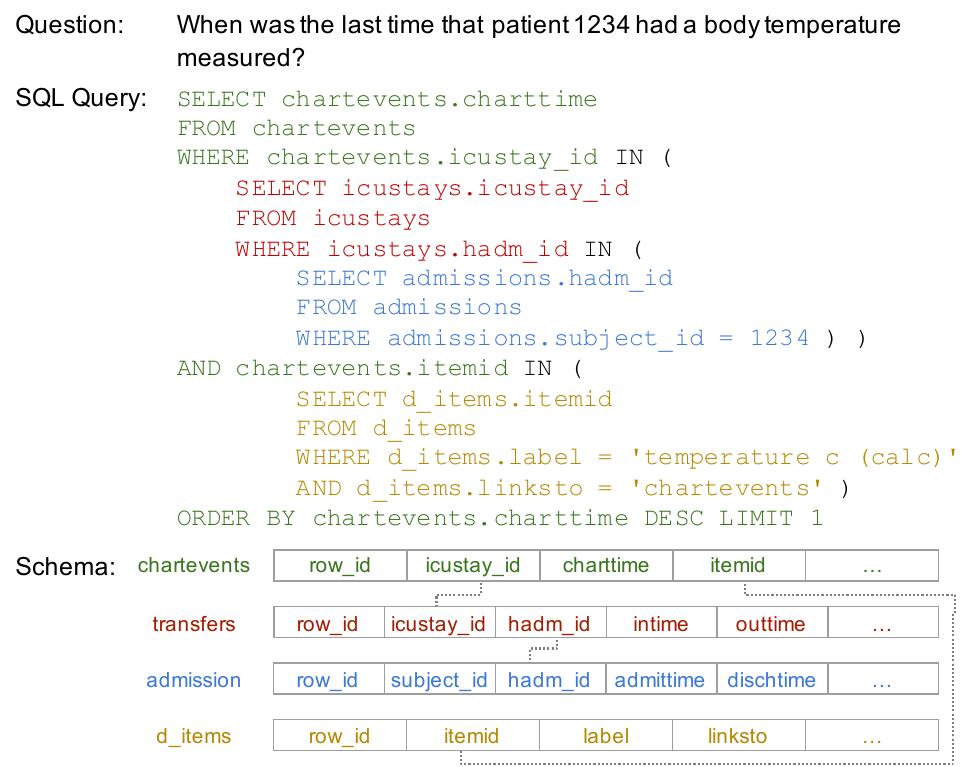}
        \caption{EHR schema is structured in a hierarchy, and real-world questions often require multi-step reasoning over tables.} 
        \label{figure:query_example}
    \end{minipage}%
\end{figure}

\paragraph{Data collection}

The motivation behind our work is to construct a dataset that reflects the actual needs of hospital staff and tackles several practical issues that can arise in real-world healthcare QA systems (see Table~\ref{tab:sample-table}). To this end, we collaborated with the Konyang University Hospital\footnote{\url{https://www.kyuh.ac.kr/eng/}} and conducted a poll to collect real-world questions as if one is asking an AI speaker what they frequently look for in structured information in the EHR. In addition, to clarify what machines can and cannot answer, we provided typical negative examples of what should not be asked, such as requiring external knowledge, ambiguous or qualitative statements, or asking for reasons behind clinical decisions. As a result, we gathered a total of 1,742 utterances, and the number of valid respondents was 222 (see Figure~\ref{figure:data_source} for respondent demographics).

\subsection{Question and SQL Generation}

\subsubsection{Question template}
\label{ssec:question_template}

After the poll, we filtered out the utterances that did not meet the criteria, including ambiguous statements, those that require external knowledge, or those that go beyond the database schema. We then categorized the utterances into three patient-based scopes: a single patient, a group of patients, and no patient.
For each scope, duplicate utterances, but differently phrased, were merged into a single question template.
We also manually added question templates that were simple variants of existing templates.
For example, we modified the type of \textit{medical events} (\textit{e.g.}, lab test, prescription, etc.) in question templates, such as from ``Count the number of times that a patient had a lab test'' to ``Count the number of times that a patient received a prescription.''
To cover more questions on longitudinal statistics, we linked two different medical events together whenever possible.
For example, the question ``What are the top \texttt{N} prescriptions after a diagnosis,'' which was originally collected, is extended to ``What are the top \texttt{N} lab tests after a diagnosis.''
Finally, the filtered-out utterances were also templatized and considered unanswerable. The unanswerable questions included side effects, next check-up schedules, primary care doctor, patient's personal information, whether a patient has signed a consent form, diagnosis received in other departments, and more.

\begin{table}[h!]
  \caption{Sample question templates. The full list is reported in Supplementary~\ref{sec:full_question_template}.}
  \small
  \centering
  \renewcommand{\arraystretch}{0.6}
  \begin{adjustbox}{width=\columnwidth,center}  
  \begin{tabular}{cl}
    \toprule
    \textbf{Patient scope} & \multicolumn{1}{c}{\textbf{Question template}} \\
    \midrule
    None                  & What is the intake method of \texttt{\{drug\_name\}}? \\
    \midrule
    Individual             & Has patient  \texttt{\{patient\_id\}} received any procedure \texttt{[time\_filter\_global]}? \\
    \midrule
    \multirow{3.3}{*}{Group} & Count the number of hospital visits of patient \texttt{\{patient\_id\}} \texttt{[time\_filter\_global]}. \\
    \cmidrule(r){2-2}
                          & \makecell[l]{What are the top \texttt{[n\_rank]} frequent procedures that patients received \texttt{[time\_filter\_within]} \\ after having been diagnosed with \texttt{\{{diagnosis\_name\}}} \texttt{[time\_filter\_global]}? }\\
    \midrule
    \makecell{Unanswerable} & What is the effect of \texttt{\{drug\_name\}}? \\
    \bottomrule
  \end{tabular}
  \end{adjustbox}  
  \label{tab:question_template_sample}
\end{table}

The resulting question templates are natural utterances with \textit{slots} that are later filled with pre-defined values or database records (see Table~\ref{tab:question_template_sample}).
Slots containing ``time\_filter'' are processed in the time template sampling stage, described in Section~\ref{ssec:time_template}.
We eliminate any ambiguity in the question templates that could cause the model to infer table or column names incorrectly.
For example, ``When did this patient get [slot] today?'' is removed because it does not tell us whether we are interested in drugs, lab tests, or something else unless the slot is filled.
Such linguistic ambiguity is later introduced in the paraphrase generation stage described in Section~\ref{ssec:paraphrase_generation}.
As a result, we curated 230 question templates (174 for answerable and 56 for unanswerable) based on the MIMIC-III and eICU schemas.
For answerable questions, their corresponding SQL queries were labeled for each database, which is further discussed in Section~\ref{ssec:sql_annotation}.

\subsubsection{Time template}
\label{ssec:time_template}

Since the healthcare domain is inextricably linked to time, many utterances we collected were rich in time expressions. To reflect this in the dataset, we developed three time filter types and assigned different time factors (expression types, units, and interval types) to each filter type to compose a single time template.
First, we define a ``global'' time filter (\texttt{[time\_filter\_global]}) that constrains the full range of time we are interested in.
Second, within the global time filter, we can also use a ``within'' time filter (\texttt{[time\_filter\_within]}) to indicate the time range between two or more medical events.
Third, we can point to an exact time (\texttt{[time\_filter\_exact]}), such as ``last'' measurement or ``at 2105-12-31 09:00:00'' if we know the exact time or the order of an event.

Each time filter type has three factors and an extra option: 
1) \textit{Expression type} is whether the filter is phrased in an absolute, relative, or mixed time expression (\textit{e.g.}, last year, in 2022, etc.);
2) \textit{Unit} is the granularity of time, such as a standardized unit (\textit{e.g.}, year, month, day, etc.) or an arbitrary unit of events (\textit{e.g.}, hospital visit, ICU visit);
3) \textit{Interval type} specifies the type of time interval (\textit{e.g.}, since, until, in, etc.). Depending on the use case, \textit{Option} allows to choose one exact event (\textit{e.g.}, first, last) among the filtered events. Table~\ref{tab:time_expression_sample} shows sample time templates and their natural language (NL) \textit{time expressions}.
It is important to note that each time template has its corresponding NL time expression (\textit{e.g.}, since \texttt{\{month\}}/\texttt{\{day\}}/\texttt{\{year\}}) and \textit{SQL time pattern} (\textit{e.g.}, \texttt{WHERE strftime(\textquotesingle\%Y-\%m-\%d\textquotesingle,[time\_column]) >= \textquotesingle\{year\}-\{month\}-\{day\}\textquotesingle)}, while the question templates are originally created in natural language.
The NL time expressions are combined with the question templates to form complete questions. The SQL time patterns are combined with the labeled SQL queries to form complete SQL queries (see Section~\ref{ssec:question_generation} for details).
The full list of time templates (NL and SQL pairs) is reported in Supplementary~\ref{sec:full_time_template}.

\begin{table}[h!]
  \caption{Sample time templates and their NL time expressions.}
  \small
  \centering
  \renewcommand{\arraystretch}{1.2}
  \begin{adjustbox}{width=\columnwidth,center}  
  \begin{tabular}{ccccccl}
    \toprule
    \textbf{Time filter type} & \textbf{Expression type} & \textbf{Unit} & \textbf{Interval type} & \textbf{Option} & \textbf{Time template} & \multicolumn{1}{c}{\textbf{NL time expression}} \\
    \midrule
    \multirow{4.2}{*}{\texttt{[time\_filter\_global]}} & absolute & day & since & - & abs-day-since & since \texttt{\{month\}}/\texttt{\{day\}}/\texttt{\{year\}} \\
    \cmidrule(r){2-7} 
                                            & \multirow{2.5}{*}{relative} & \multirow{2.5}{*}{year} &  \multirow{2.5}{*}{in}  & last & rel-year-last & last year \\
    \cmidrule(r){5-7}
                                            & & & & this & rel-year-this & this year \\
    \midrule
    \multirow{2.6}{*}{\texttt{[time\_filter\_within]}} & - & hospital & in & -  & within-hosp & within the same hospital visit \\
    \cmidrule(r){2-7}
                                            & - & day & in  & - & within-day & within the same day \\    
    \midrule    
    \texttt{[time\_filter\_exact]} & relative & exact & at & - & exact-first & first \\
    \bottomrule
  \end{tabular}
  \end{adjustbox}  
  \label{tab:time_expression_sample}
\end{table}

\subsubsection{SQL annotation}
\label{ssec:sql_annotation}

The SQL annotation process was conducted manually by four graduate students over the course of five months, with repeated revisions.
Since the questions were collected independently of the database schema, SQL labeling required numerous assumptions (\textit{e.g.}, which medical event is mapped to which column in the database, the choice of rank functions in SQL, the occasion of using \texttt{DISTINCT}, etc.).
To sync template and value sampling, the same slots introduced in Section~\ref{ssec:question_template} and~\ref{ssec:time_template} (\textit{e.g.}, \texttt{\{drug\_name\}}, \texttt{[n\_rank]}, \texttt{[time\_filter\_global]}) were also used in SQL labeling as placeholders.
The SQL queries for question and time templates were labeled by one person, followed by a reviewer for each database.
Unlike most SQL datasets, the students were asked to avoid using \texttt{JOIN}, which is a go-to operation when retrieving information across multiple tables.
In MIMIC-III and eICU, several tables contain more than 100 million rows (\textit{e.g.}, 330 million rows in MIMIC-III chartevents) as the records are stored in a ``log'' based manner.
In fact, it is extremely inefficient to merge all the tables in such large databases without understanding what specific columns are needed to answer a question. As a result, the students were asked to utilize the hierarchical structure of the EHR schema, as illustrated in Figure~\ref{figure:query_example}, and labeled each query in a nested manner whenever possible. SQL annotation details and the comparison between \texttt{JOIN} and nesting-based queries are discussed in Supplementary~\ref{sec:sql_labeling_detail}.

\begin{figure}[t!]
\centering
\includegraphics[width=\linewidth]{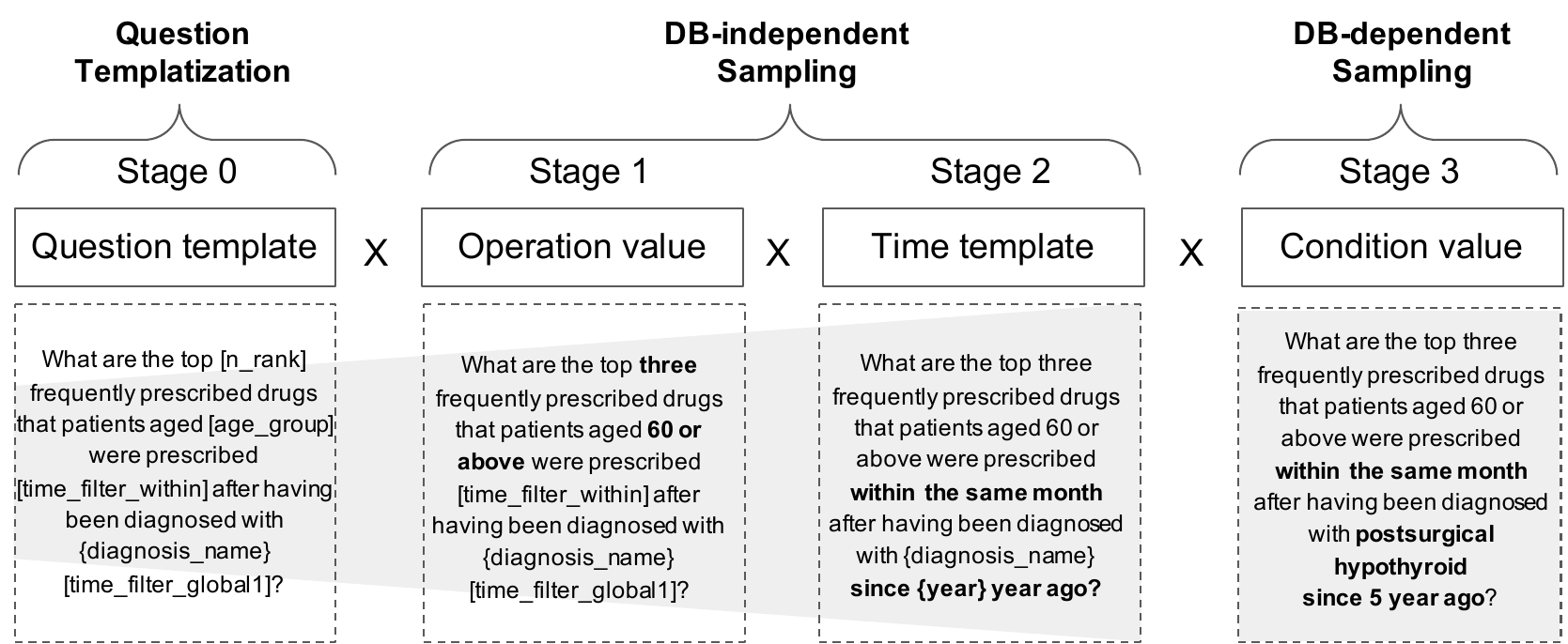}
\caption{Question generation process. The shaded region refers to the expansion of semantic diversity in the template pool.}
\label{figure:question_generation}
\end{figure}

\subsubsection{Template combination and data generation}
\label{ssec:question_generation}

Data generation starts by selecting a question template, followed by three sampling stages that add semantic variety: \textit{operation value sampling}, \textit{time template sampling}, and \textit{condition value sampling} (See Figure~\ref{figure:question_generation}). Stage 1 involves sampling operation values from pre-defined values that are independent of the database schema or records (\textit{e.g.}, maximum, five-year, two or more times, etc.). Then, time templates are sampled in Stage 2 based on the time filter types that each question template can have. Stage 2 in Figure~\ref{figure:question_generation} illustrates that the time templates (\texttt{[time\_filter\_global]} and \texttt{[time\_filter\_within]}) have already been sampled and converted into NL time expressions. More details are discussed in Supplementary~\ref{sec:template_combination}.
Finally, condition value sampling (\textit{e.g.}, \texttt{\{diagnosis\_name\}}, \texttt{\{year\}}) occurs in Stage 3, and the slot filling process is complete. This final stage is where the actual database records are sampled, and all generated SQL queries are checked to have at least one valid answer. Once the query is successfully executed, the pair of SQL and the corresponding question is added to the data pool, which is later used for data splitting.

\subsection{Database Pre-processing}
\label{ssec:database_preprocessing}

\paragraph{Cost table} 
We modified the original MIMIC-III and eICU databases to include a cost table in order to reflect hospital administration and insurance-related questions.
When constructing the table, we referred to the OMOP Common Data Model\footnote{\url{http://ohdsi.github.io/CommonDataModel/cdm54.html}} to create new column names, which include: \texttt{~patient\_id},\texttt{~hospital\_admission\_id},\texttt{~event\_type} (cost\_domain\_id),\texttt{~event\_id} (cost\_event\_id),\texttt{~chargetime}, and\texttt{~cost}.
For cost sampling, we take a two-step sampling procedure to simulate cost values. First, we sampled discrete-valued mean costs for four different medical event types (diagnosis, procedure, prescription, and lab events) from a Poisson distribution with a mean of 10. Then, we sampled continuous-valued costs from Gaussian distributions with their corresponding means.

\paragraph{Time shifting}
The time span of MIMIC-III is over a hundred years due to the de-identification process (originally seven years), while eICU's time span remains intact (two years). 
To simulate a more realistic time span asked in the hospital and match the time across the databases, we shifted the admission time of every patient's records ranging from 2100 to 2105.
In addition, to incorporate the concept of \textit{current} and the use of relative time expressions, we set the current time to be \texttt{``2105-12-31 23:59:00''} and removed any record past the current time. In this way, patients with missing hospital discharge time are considered current patients. The poll result revealed that many time expressions used in the hospital are relative time expressions (\textit{e.g.}, today, yesterday, last month, etc.), and therefore this process was necessary to reflect time-sensitive questions. More details are reported in Supplementary~\ref{sec:time_shifting}.

\paragraph{De-identification}
MIMIC-III and eICU databases are de-identified datasets, and a user needs to request credentialed access to PhysioNet\footnote{{\url{https://physionet.org/}}} to obtain them. Both datasets, however, do contain real patient records, and the questions derived from them could potentially reveal patient-specific information. For example, the question ``What medication was given to patient ID 1234 after being diagnosed with diabetes?'' implies that this patient was diagnosed with diabetes. Combined with some unforeseen external knowledge, this might lead to recovering the patient's identity. To add another layer of de-identification, we randomly shuffled values across all patients in the database before sampling condition values. In this way, the semantic structure of the question remains the same, but sampled condition values are untraceable. More details of the de-identification process are reported in Supplementary~\ref{sec:deidentification}.

\begin{table}[t!]
  \caption{The comparison between EHRSQL and other text-to-SQL datasets. \textit{\# Nesting/Q} is the average number of nesting levels per query. \textit{\% Time Used/Q} is the percentage of at least one time column used in queries. \textit{?Schema} is whether the question authors are unaware of the database schema when they first come up with their question. \textit{UnANS} indicates whether the dataset contains unanswerable questions.} 
  \small
  \centering
  \renewcommand{\arraystretch}{1.0}
  \begin{adjustbox}{width=\columnwidth,center}  
  \begin{tabular}{cccccccccc}
    \toprule
    \textbf{Dataset} & \textbf{\# Example} & \textbf{\# DB} & \textbf{\# Table/DB} & \textbf{\# Row/Table} & \textbf{\# Table/Q} & \textbf{\# Nesting/Q} & \textbf{\% Time Used/Q} & \textbf{?Schema} & \textbf{UnANS} \\
    \midrule
    Spider\textsuperscript{$\dagger$}   & 8K   & 160   & 5.1  & 2K     & 1.6  & 1.2 & 12.7\% & \xmark & \xmark \\
    KaggleDBQA      & 0.3K & 8     & 2.3  & 280K   & 1.2    & 1.0   & 17.3\%     & \xmark & \xmark \\    
    SEDE            & 12K  & 1     & 29   & -      & 1.8  & 1.3 & 20.2\%   & \xmark & \xmark \\    
    \midrule
    MIMICSQL        & 10K  & 1     & 5    & 7K     & 1.8    & 1.0 & 26.6\%     & \xmark & \xmark \\
    emrKBQA\textsuperscript{$\ddagger$} & 940K & 1     & 9    & -      & -    & -   & -     & \cmark & \xmark \\
    \midrule
    \textbf{EHRSQL} & 24K  & 2     & 13.5 & 108K   & 2.4  & 2.7 & 93.2\% & \cmark & \cmark \\    
    \bottomrule
    \multicolumn{10}{l}{{$\dagger$}Train and validation sets are counted. {$\ddagger$}The dataset is yet to be publicly released.}
    \vspace{-5mm}
  \end{tabular}
  \end{adjustbox}  
  \label{tab:dataset_comparison}
\end{table}

\subsection{Paraphrase Generation}
\label{ssec:paraphrase_generation}

To add linguistic variety to the questions, we generated template paraphrases from the question templates with manual paraphrasing and the help of machine learning models. Before paraphrasing, we ensure that the templates do not contain any domain-specific expressions, making the paraphrasing process depart from the healthcare domain to leverage general-domain tools. The overall procedure is as follows:

\begin{enumerate}[leftmargin=5.5mm]
  \item Human paraphrasing is first conducted to add more high-quality templates, averaging 21 paraphrases per question template.
  \item Based on the human paraphrases, more paraphrases are generated using machine learning models, such as T5 paraphrasers~\cite{prithivida2021styleformer, prithivida2021parrot, alisetti2021t5} and multilingual translation models for back-translation~\cite{fan2021beyond, tang2020multilingual}.
  \item The paraphrases that are too different from the original meaning are filtered using a duplicate question detection model, specifically RoBERTa-large~\cite{liu2019roberta} trained on the Quora duplicate question detection dataset.
  \item The paraphrases are ranked by the perplexity score from GPT-Neo 1.3B~\cite{gao2020pile} per question template and disregarded if they are too similar to the other paraphrases with lower perplexity. The Levenshtein distance is used to detect lexical similarity.
  \item The final paraphrases are reviewed by crowd workers to rate the quality of the paraphrases. In our case, we collaborated with a crowd-sourcing company named Selectstar\footnote{\url{https://selectstar.ai/en}}. Three groups of annotators were assigned to this task and marked a pass or fail for each paraphrase sample. The samples with unanimous pass marks were considered the final machine-paraphrased templates, leaving 47 paraphrases per template on average. 
\end{enumerate}

During the paraphrasing process, slots are replaced with generic values (\textit{e.g.}, \texttt{\{patient\_id\}} $\rightarrow$ the patient) and returned to the original slots after the process is complete. Later, those slots are filled with the actual condition values in the value sampling process to construct the final question-to-SQL pairs. The overview of the paraphrase pipeline is illustrated in Supplementary~\ref{sec:template_paraphrasing}.

\subsection{EHRSQL and Other Datasets}

Table~\ref{tab:dataset_comparison} summarizes the statistics of EHRSQL and other text-to-SQL datasets in the literature. Compared to general domain datasets (first three rows), EHRSQL has a large number of text-to-SQL pairs (\textit{\# Example}), tables per database (\textit{\# Table/DB}), and rows per table (\textit{\# Row/Table}). KaggleDBQA~\cite{lee2021kaggledbqa} and SEDE~\cite{hazoom2021text} are designed to bridge the gap between academic datasets and practical usability by using real databases and naturally-occurring utterances. However, we have gone one step further where the question authors (the poll respondents) were not presented with the database schema (\textit{?Schema}), which adds more reality to the dataset~\cite{hazoom2021text}. Finally, EHRSQL contains unanswerable questions (\textit{UnANS}) that were collected together from the poll, which may play a critical role in assessing the reliability of the QA model. To the best of our knowledge, EHRSQL is the first attempt to collect and combine answerable and unanswerable questions in the context of text-to-SQL.

From a healthcare perspective, EHRSQL covers a variety of questions frequently asked in the hospital. The scope of these questions ranges from non-patient information (\textit{e.g.}, the cost of a procedure) to individual-level information (\textit{e.g.}, retrieving patient demographics and the lab values) and group-level information (\textit{e.g.}, counting the number of patients in the hospital and the top \texttt{N} frequently prescribed drugs of patients over 60s), further combined with various time expressions (\textit{\% Time Used/Q}). The SQL queries are linked to two EHR databases: MIMIC-III and eICU, and this is the first attempt to label SQL queries on eICU, allowing questions about multi-center critical care to be answered.

\section{Benchmarks}
\subsection{Task}
\label{ssec:task}

\begin{figure}[t!]
\centering
\includegraphics[width=\linewidth]{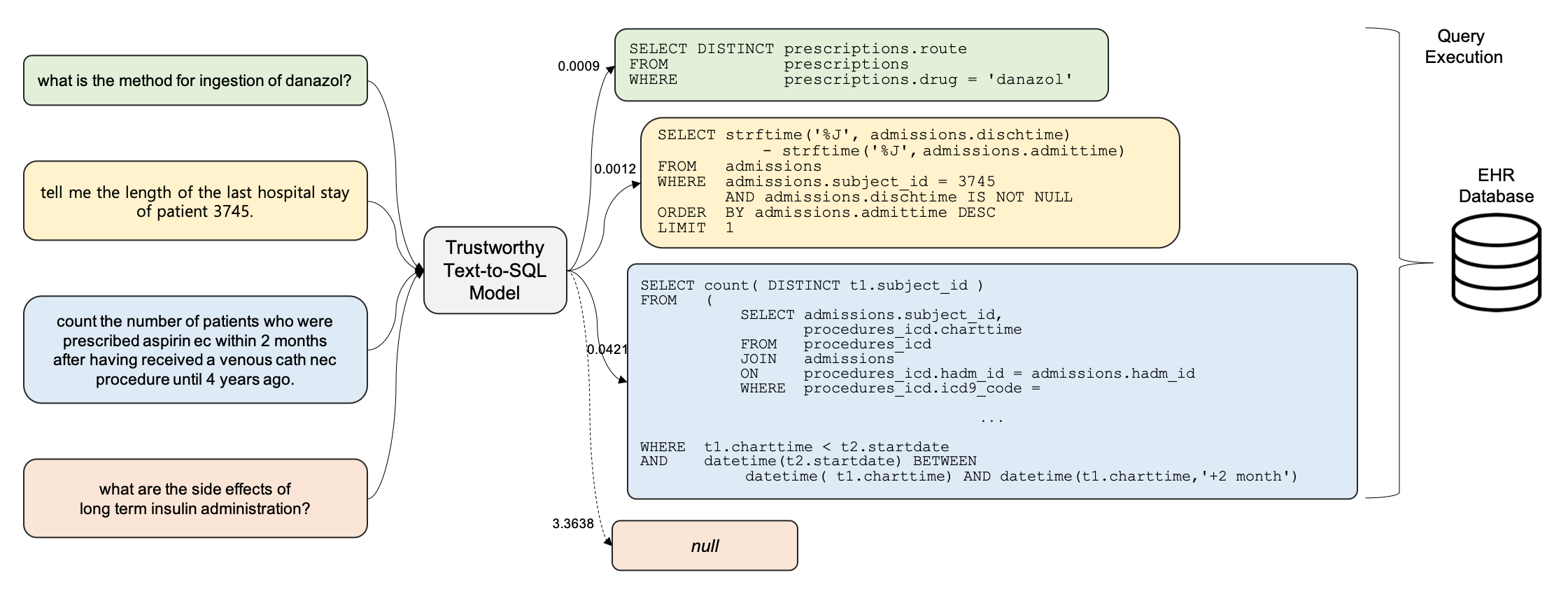}
\caption{Overview of trustworthy semantic parsing.}
\label{figure:task_overview}
\end{figure}

In this section, we define a new text-to-SQL task that assesses models under more realistic healthcare QA settings than prior works—namely, trustworthy semantic parsing. As illustrated in Figure~\ref{figure:task_overview}, the trustworthy model needs to 1) generate SQL queries that reflect a wide range of needs in the hospital workplace, 2) understand various time expressions to answer time-sensitive questions in healthcare, and 3) have the capacity to distinguish whether a given question is answerable or unanswerable based on the prediction confidence. Among various ways to calculate the confidence score, we argue that a desirable way to calculate it is not based on the data the model is trained on, but on the model prediction itself. As illustrated in Figure~\ref{figure:task_overview}, once a confidence score exceeds some threshold of choice (assuming that a greater score means less confidence), the generated SQL should not be sent to the database.

To make the task feasible, we keep the condition values (\textit{e.g.}, prescription name) the same in this work, except for genders and vital signs, as this is another major challenge in semantic parsing~\cite{suhr2020exploring, shi2021learning}. 
When splitting the dataset into train, validation, and test sets, we ensure that all the question templates are present in each split. To create a more realistic QA setting, we include unanswerable questions only in the validation and test sets (consisting of 33\% of each split), following the OOD detection setting~\cite{marek2021oodgan, zheng2020out}. Among 24,411 question pairs in the dataset, the train, valid, and test splits contain 9.3K, 1.1K, and 1.8K pairs, respectively, for each database. We release the train and valid splits, totaling 21K samples, and the other 3K samples are saved for a hidden test set. Details of data splitting are reported in Supplementary~\ref{sec:data_splitting}.

\subsection{Evaluation}

The goal of the trustworthy semantic parsing task is to correctly return the answers to answerable questions while disregarding unanswerable questions. This process can be evaluated in two aspects. The model first needs to have a strategy that distinguishes whether a given question is answerable or unanswerable.
Given that the goal is to correctly recognize as many answerable questions as possible, the model's performance can be measured in precision and recall.
In this case, the precision (P$_{ans}$) calculates the number of correctly recognized answerable questions among all questions predicted to be answerable, while recall (R$_{ans}$) calculates the number of correctly recognized answerable questions among all answerable questions.
We use F1$_{ans}$ to combine these two scores.

Next, we must evaluate how well the model generates correct SQL queries, given that some questions are considered answerable and ready to predict.
The word \textit{predict} here means the actions of generating a query and sending it to the database, where extra care is needed because the answer from the database may be directly used for clinical decision-making.
Combining the concept of F1$_{ans}$ and execution accuracy in semantic parsing, F1$_{exe}$ counts only when the returned answer is correct.
In other words, F1$_{exe}$ is a combined score of P$_{exe}$ and R$_{exe}$, where P$_{exe}$ is the ratio of the number of correctly answered questions to all questions predicted to be answerable and R$_{exe}$ is the ratio of the number of correctly answered questions to all answerable questions. 
As a result, the denominators of the precisions and recalls are the same, but the numerators are more strictly counted in metrics with ${exe}$.
This gap between ${ans}$ and ${exe}$ reveals the model's inability to generate correct SQL queries, given that the questions are both recognized as answerable and indeed answerable.

\begin{table}[!t]
\renewcommand{\arraystretch}{1.1}
    \centering
    \caption{Performance on MIMIC-III (Left) and eICU (Right).}
    \label{tab:main_result}
    \begin{minipage}{.49\textwidth}
      \centering
        \scalebox{0.58}{
        \begin{tabular}{lcccccccc}
        \toprule
          \multicolumn{1}{c}{\multirow{2.4}{*}{\textbf{Model}}} & \multicolumn{4}{c}{\textbf{Valid}} & \multicolumn{4}{c}{\textbf{Test}} \\
          \cmidrule(lr){2-5}
          \cmidrule(lr){6-9}
          & F1$_{ans}$ & P$_{exe}$ & R$_{exe}$ & F1$_{exe}$ & F1$_{ans}$ & P$_{exe}$ & R$_{exe}$ & F1$_{exe}$ \\
          \midrule
          & \multicolumn{8}{c}{\textit{Threshold: None}} \\
          T5           & 80.8 & 65.2 & 96.3 & 77.8 & 80.3  & 64.0 & 95.4 & 76.6 \\ 
          T5 + Schema  & 80.8 & 65.3 & 96.4 & 77.9 & 80.3  & 64.0 & 95.3 & 76.6 \\   
          \midrule
          & \multicolumn{8}{c}{\textit{Threshold: Clustering-based}} \\
          T5           & 89.2 & 80.2 & 95.5 & 87.2 & 86.5 & 75.0 & 94.8 & 83.8 \\ 
          T5 + Schema  & 87.9 & 77.3 & 96.2 & 85.7 & 84.8 & 72.3 & 94.9 & 82.1 \\
          \midrule
          & \multicolumn{8}{c}{\textit{Threshold: Percentile-based}} \\
          T5           & 94.8 & 94.3 & 93.3 & 93.8 & 91.4  & 88.8 & 91.1 & 89.9 \\ 
          T5 + Schema  & 93.5 & 93.0 & 92.0 & 92.5 & 90.6  & 88.4 & 89.7 & 89.1 \\     
          \bottomrule
        \end{tabular}
        }
    \end{minipage}
    \begin{minipage}{.49\textwidth}
      \centering
        \scalebox{0.58}{
        \begin{tabular}{lcccccccc}
        \toprule
          \multicolumn{1}{c}{\multirow{2.4}{*}{\textbf{Model}}} & \multicolumn{4}{c}{\textbf{Valid}} & \multicolumn{4}{c}{\textbf{Test}} \\
          \cmidrule(lr){2-5}
          \cmidrule(lr){6-9}
          & F1$_{ans}$ & P$_{exe}$ & R$_{exe}$ & F1$_{exe}$ & F1$_{ans}$ & P$_{exe}$ & R$_{exe}$ & F1$_{exe}$ \\
          \midrule
          & \multicolumn{8}{c}{\textit{Threshold: None}} \\
          T5           & 80.7 & 65.2 & 96.4 & 77.8 & 80.4  & 64.0 & 95.2 & 76.5 \\ 
          T5 + Schema  & 80.7 & 64.4 & 95.2 & 76.8 & 80.4  & 64.1 & 95.4 & 76.7 \\   
          \midrule
          & \multicolumn{8}{c}{\textit{Threshold: Clustering-based}} \\
          T5           & 88.4 & 78.4 & 95.5 & 86.1 & 86.1  & 73.2 & 94.5 & 82.5 \\ 
          T5 + Schema  & 86.6 & 73.8 & 95.1 & 83.1 & 85.1  & 71.4 & 95.4 & 81.7 \\
          \midrule
          & \multicolumn{8}{c}{\textit{Threshold: Percentile-based}} \\
          T5           & 93.0 & 92.5 & 91.7 & 92.1 & 90.8  & 87.5 & 91.6 & 89.5 \\ 
          T5 + Schema  & 94.2 & 92.5 & 91.7 & 92.1 & 90.9  & 84.8 & 93.5 & 88.9 \\
          \bottomrule
        \end{tabular}
        }
    \end{minipage}
  \end{table}

\subsection{Model Development}
Most state-of-the-art (SOTA) semantic parsing models utilize grammar-based decoders and are only compatible with the grammars (or their subsets) defined in the Spider dataset. As the queries in EHRSQL are created in response to actual needs, most of them are incompatible with the Spider parser mainly due to time-related operators like \texttt{~strftime},\texttt{~datetime} and\texttt{~NULL}.
Therefore, we chose general-purpose sequence-to-sequence models, T5-base~\cite{raffel2019exploring} and T5-base with schema serialization~\cite{suhr2020exploring, hazoom2021text}, as baseline models for our task. This choice aligns with a recent finding that transfer learning from pre-trained language models surpasses healthcare-specific text-to-SQL models~\cite{pmlr-v158-bae21a}. Training details are reported in Supplementary~\ref{sec:training_detail}.

To give a model the ability to refuse to answer questions, we adopt a simple uncertainty estimation method inspired by \cite{malinin2020uncertainty, xiao2021hallucination}.
If the maximum entropy during the decoding process exceeds a pre-defined threshold, we consider this a refusal. The threshold values are determined by two heuristic approaches, and they are compared with the result obtained without refusal.

\begin{enumerate}[leftmargin=5.5mm]
  \item Clustering-based: Run K-means clustering with $k=2$ on all maximum entropy values of validation samples, assuming that high-entropy samples are from unanswerable questions. The decision boundary between the two clusters is then used as the threshold.  
  \item Percentile-based: Set the threshold to the 67th percentile of the maximum entropy values of the validation samples, based on the fact that 33\% of the questions in the validation set are unanswerable.
\end{enumerate}

To illustrate the domain gap between general-domain and healthcare datasets, we create a subset of EHRSQL that could be parsed with the Spider parser. Then, we compare how well the SOTA cross-domain semantic parsing models generalize to healthcare text-to-SQL datasets. We choose MIMICSQL, which contains simple SQL queries that can all be parsed with the Spider grammar, and EHRSQL for the healthcare datasets. Among many SOTA models in the Spider leaderboard, we use Generation-Augmented Pre-training (GAP)~\cite{shi2021learning} to test its zero-shot domain transfer performance.

\subsection{Results and Findings}

The baseline results are reported in Table~\ref{tab:main_result}.
Among the three different threshold approaches, the percentile-based threshold performs best on both the validation and test sets.
This result is an expected outcome because the ratios of the answerable and unanswerable questions in the validation and test sets are kept the same in our setting.
As for the clustering-based method, an additional training or clustering technique is needed for the models to better discriminate the entropy values. The naive entropy values from the current models are not linearly separable across predictions (see Figure 7 in Supplementary~\ref{sec:result_entropy} for maximum entropy distributions).
T5+Schema shows comparable performance to T5 in both databases. This result agrees with the recent finding in \cite{hazoom2021text} that models trained in a single database setting do not effectively leverage schema information.
Additional qualitative results are provided in Supplementary~\ref{sec:qualitative_result}, including SQL generation results by question complexity, time expressions, falsely executed results, and refused results.

\begin{table}[t!]
  \caption{Performance of zero-shot cross-domain transfer with GAP in execution accuracy. Easy, medium, hard, and extra refer to Spider's SQL hardness criteria. The numbers in parenthesis indicate the number of correct cases divided by the number of answerable queries. All experiments are done on answerable questions in the validation set.}
  \small
  \centering
 \scalebox{0.755}{  
  \renewcommand{\arraystretch}{1.1}
  \begin{tabular}{lcccccc}
    \toprule
    \multicolumn{1}{c}{\textbf{Dataset}} & \textbf{Easy}  & \textbf{Medium} & \textbf{Hard}  & \textbf{Extra} & \textbf{Skipped} & \textbf{All $\textbackslash$ Skipped} \\
    \cmidrule(lr){1-1}
    \cmidrule(lr){2-6}
    \cmidrule(lr){7-7}
    MIMICSQL\cite{wang2020text} & 46.7 (79/169)  & 10.5 (85/809)  & 0.0 (0/22) & - & - & 16.4 (164/1000) \\ 
    \cmidrule(lr){1-1}
    \cmidrule(lr){2-6}
    \cmidrule(lr){7-7}    
    EHRSQL      & 23.8 (5/21) & - & 0.0 (0/23) & 0.0 (0/63)   & (0/1408) & 4.7 (5/107) \\
    \quad MIMIC-III only & 16.7 (2/12) & - & 0.0 (0/15) & 0.0 (0/30)   & (0/703)  & 3.5 (2/57) \\
    \quad eICU only      & 3.3 (3/9) & - & 0.0 (0/8) & 0.0 (0/33)   & (0/705)  & 6.0 (3/50) \\
    \bottomrule
  \end{tabular}
  }
  \label{tab:cross_domain_result}
\end{table}

Table~\ref{tab:cross_domain_result} shows the performance of zero-shot cross-domain transfer with the GAP model. Unlike the queries in MIMICSQL, which are all parsable with the Spider grammar, EHRSQL has much more complex SQL operators and structures, with only about 7\% (107/1,515) of the answerable questions in the validation set being parsable. As for model performance, GAP achieves 16.4\% in the full MIMICSQL validation set and 4.7\% in the subset of the validation set in EHRSQL. This result motivates again the need for a new practical text-to-SQL dataset in healthcare and QA models that can handle multiple real-world challenges in the hospital.

\section{Conclusion and Future Direction}
In this paper, we present EHRSQL, a new practical text-to-SQL dataset for question answering over structured information in EHRs. Through a poll conducted at a university hospital, we collected questions that are frequently asked on structured EHR data across various professions in the hospital. The questions reflect the actual needs in the hospital and different time expressions used in daily work, which are particularly crucial in healthcare. Additionally, we also collected unanswerable questions, which were the questions submitted by the respondents but turned out to be beyond the EHR schema or ambiguous. Finally, we manually labeled SQL queries for two open-source EHR databases—MIMIC-III and eICU—and cast these challenges as one task—trustworthy semantic parsing—where QA models should only answer the question when their predictions are confident but not otherwise.

Though we have carefully designed the dataset, there are several limitations. First, despite more than two hundred people participating in the poll, the source of the questions is from one Korean university hospital, which may not reflect every unique situation in different hospitals worldwide. Secondly, the question templates are paraphrased using general-domain paraphrasers; therefore, paraphrases with the heavy use of medical jargon could make them more realistic. Finally, SQL labels for two EHR databases might not be a sufficient number of databases to train and test a model for unseen EHR databases.

We expect numerous research directions with our dataset. The seed questions can be a valuable resource for expanding the scope of table-based healthcare QA tasks, such as creating interactive QA~\cite{yu2019sparc, elgohary-etal-2020-speak} or developing it into multimodal QA datasets on EHRs. With the idea of trustworthy semantic parsing, a new class of end-to-end uncertainty-aware semantic parsing models can be proposed. As existing text-to-SQL models assume all inputs are answerable, this can be a unique venue for semantic parsing to bridge the gap between research and industrial needs.

\clearpage

\begin{ack}
We would like to thank five anonymous reviewers for their time and insightful comments. We also thank Woochan Hwang for assisting us in clarifying our questions during the data labeling process. This work was supported by Institute of Information \& communications Technology Planning \& Evaluation (IITP) grant funded by the Korea government(MSIT) (No.2019-0-00075, Artificial Intelligence Graduate School Program(KAIST)), National Research Foundation of Korea (NRF) grant (NRF-2020H1D3A2A03100945) and Data Voucher grant (2021-DV-I-P-00114), funded by the Korea government (MSIT).
\end{ack}


{
\small
\bibliographystyle{plainnat}
\bibliography{references}
}

\appendix
\clearpage
\include{supplementary}

\end{document}

%% file: supplementary.tex
\section*{Supplementary Material}

\section{Datasheet for Datasets}
\label{sec:datasheet_for_datasets}

The following section is answers to questions listed in datasheets for datasets.

\input{supp/datasheet_for_dataset}

\section{Full List of Templates}

\subsection{Question Templates}
\label{sec:full_question_template}

The full list of question templates is reported in Table~\ref{tab:question_template_full}. The total number of answerable question templates is 174, but a few can be unanswerable depending on the database (\textit{e.g.}, a question about a procedure done in other hospitals is not answerable in MIMIC-III). For unanswerable questions, the template generation process explained in Section 3.1.1 is not strictly applied. As a result, unanswerable questions can contain ambiguous and too detailed questions (\textit{e.g.}, Tell me what medicine to use to relieve a headache in hypertensive patients). We do not provide a full list of unanswerable questions as they are not subject to training and can be anything complementary to the answerable ones.

\input{supp/question_template}


Several question templates assume a specific range of patients (\textit{e.g.}, current patients or already discharged patients), as indicated in the ``Assumption'' column in Table~\ref{tab:question_template_full}. For example, depending on the patients, the way to calculate the duration of hospital stay can be different. The SQL query for already discharged patients (\textit{i.e.}, What was the \texttt{[time\_filter\_exact1]} length of hospital stay of patient \texttt{{patient\_id}}?) calculates the time between the hospital admission and discharge, while the same query for the current patients (\textit{i.e.}, How many \texttt{[unit\_count]} have passed since patient \texttt{{patient\_id}} was admitted to the hospital currently?) calculates the time between the hospital admission and current time. We intentionally separated the templates to make the model better understand the hidden assumptions behind user utterances.

Time slots with numbering (\textit{e.g.}, \texttt{[time\_filter\_global1]} and \texttt{[time\_filter\_global2]} ) are the variants of the time slot without numbering (\textit{e.g.}, \texttt{[time\_filter\_global]}). 
The purpose of the numbering is to indicate the temporal order of time filters. Specifically, a higher number indicates the same or a later time for \texttt{[time\_filter\_global1]} and \texttt{[time\_filter\_global2]}, and \texttt{[time\_filter\_exact2]} must be later than \texttt{[time\_filter\_exact1]}. 

Some verbs that end with ``\_verb'' in question templates indicate that the verb tense can change depending on the sampled time templates.

\subsection{Time Templates}
\label{sec:full_time_template}

Table~\ref{tab:time_template_full} shows the full list of time templates with natural language (NL) time expressions and SQL time patterns. Based on the time filter types present in question templates, time templates are sampled, and their corresponding NL time expressions and SQL time patterns are added to the question templates and SQL queries. Column slots in the SQL time patterns such as \texttt{[time\_column]} and \texttt{[hospital\_dischargetime]} are replaced with the actual column names following the database schema. The blanks for the NL time expressions and SQL time patterns in the table indicate that no time filter is applied.

\input{supp/time_template}

Time templates with the option ``this'' are not combined with the interval type of ``since'' or ``until'' because combining ``this'' with ``since'' is equivalent to combining ``this'' with ``in'' (\textit{i.e.}, since this year is equivalent to this year). Additionally, combining ``this'' with ``until'' is equivalent to no time constraint (\textit{i.e.}, until this year is equivalent to no time filter).

\begin{figure}[h!]
\centering
\includegraphics[width=\linewidth]{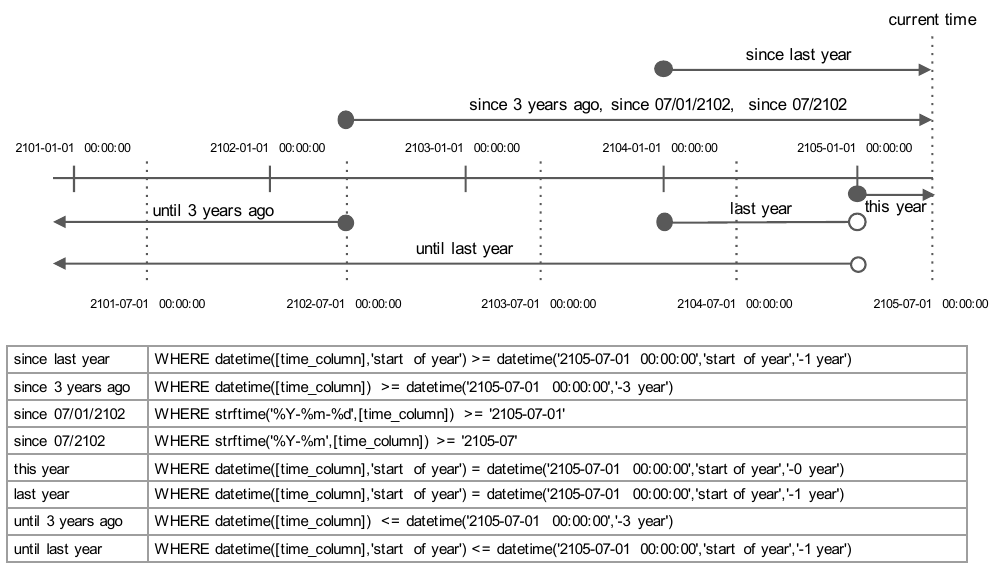}
\caption{Illustration of time templates.}
\label{figure:time_diagram}
\end{figure}

In relative time expressions, the concept of \texttt{N} units before the current time is ambiguous because one year ago and the last year can differ depending on the context. Therefore, we strictly define ``\texttt{N} units ago'' as a time point exactly \texttt{N} units before the current time (\textit{e.g.}, one year ago of the current time 2105-12-31 23:59:00 is 2104-12-31 23:59:00) and they can be combined with the ``until'' and ``since'' interval types. Figure~\ref{figure:time_diagram} illustrates several time templates.

\subsection{Template Combination}
\label{sec:template_combination}

As question templates can be combined with multiple slots, we tagged each question to track what time templates and pre-defined values are combined to form the final question. Each tag (\textit{Q\_tag}, \textit{O\_tag}, and \textit{T\_tag}) represents Stages 0, 1, and 2 in Figure 3. Except for \textit{Q\_tag}, which indicates the question template, \textit{O\_tag} and \textit{T\_tag} have fixed numbers of placeholders that store each sampled template and value. \textit{O\_tag} stores nine different types of operation values in a tuple: (\texttt{[age\_group]}, \texttt{[agg\_function]}, \texttt{[comparison]}, \texttt{[n\_rank]}, \texttt{[n\_survival\_period]}, \texttt{[n\_times]}, \texttt{[sort]}, \texttt{[unit\_average]}, \texttt{[unit\_count]}), sampled in Stage 1. \textit{T\_tag} stores time templates in a tuple: (\texttt{[time\_filter\_global1]}, \texttt{[time\_filter\_global2]}, \texttt{[time\_filter\_within]}, \texttt{[time\_filter\_exact1]}, \texttt{[time\_filter\_exact2]}), sampled in Stage 2.

A full list of the operation values is reported in Table~\ref{tab:operation_value}. Similar to time templates, the operation values have both NL expressions and SQL patterns.

\begin{table}[h!]
\centering
\caption{Pre-defined operation values.}
\scalebox{0.7}{
\begin{tabular}{ccl}
\toprule
\textbf{Operation value type} & \multicolumn{1}{c}{\textbf{NL operation expression}} & \multicolumn{1}{c}{\textbf{SQL operation pattern}} \\
\midrule
\multirow{5}{*}{\texttt{[age\_group]}} 
    & 20\textquotesingle s & WHERE \texttt{[age\_col]} BETWEEN 20 AND 29 \\
    & 30\textquotesingle s & WHERE \texttt{[age\_col]} BETWEEN 30 AND 39 \\
    & 40\textquotesingle s & WHERE \texttt{[age\_col]} BETWEEN 40 AND 49 \\
    & 50\textquotesingle s & WHERE \texttt{[age\_col]} BETWEEN 50 AND 59 \\
    & 60 or above & WHERE \texttt{[age\_col]} >= 60 \\    
\midrule    
\multirow{3}{*}{\texttt{[age\_function]}} 
    & maximum & MAX \\
    & minimum & MIN \\
    & average & AVG \\
\midrule    
\multirow{2}{*}{\texttt{[comparison]}} 
    & greater & > \\
    & less & < \\
\midrule    
\multirow{3}{*}{\texttt{[n\_rank]}}
    & three & 3 \\
    & four & 4 \\
    & five & 5 \\    
\midrule    
\multirow{5}{*}{\texttt{[n\_survival\_period]}}
    & one year & 1 * 365 \\
    & two year & 2 * 365 \\
    & three year & 3 * 365 \\    
    & four year & 4 * 365 \\
    & five year & 5 * 365 \\    
\midrule    
\multirow{2}{*}{\texttt{[n\_times]}}
    & two times & = 2 \\
    & two or more times & >= 2 \\
\midrule    
\multirow{2}{*}{\texttt{[sort]}}
    & min & ORDER BY \texttt{[sort\_col]} ASC LIMIT 1 \\
    & max & ORDER BY \texttt{[sort\_col]} DESC LIMIT 1 \\
\midrule
\multirow{3}{*}{\texttt{[unit\_average]}}
    & yearly & GROUP BY strftime(\textquotesingle \%Y\textquotesingle,\texttt{[time\_col]}) \\
    & monthly & GROUP BY strftime(\textquotesingle \%Y-\%m\textquotesingle,\texttt{[time\_col]}) \\
    & daily & GROUP BY strftime(\textquotesingle\%Y-\%m-\%d\textquotesingle,\texttt{[time\_col]}) \\
\midrule    
\multirow{2}{*}{\texttt{[unit\_count]}}
    & days & 1 * \\
    & hours & 24 * \\
\bottomrule
\end{tabular}
}
\label{tab:operation_value}
\end{table}

\section{SQL Labeling Details}
\label{sec:sql_labeling_detail}

Since the questions were collected independently of the database schema, our SQL labeling process required us to make numerous assumptions. Below is a list of the assumptions we made to label the queries.

\subsection{Shared Assumptions}
\label{sec:sql_labeling_general}

\begin{itemize}
  \item The age of a patient is calculated only once at each hospital admission time. Therefore, even if a patient stays more than a year without hospital discharge, the age remains the same.
  \item To count the number of patients or hospital (or ICU) visits, \texttt{DISTINCT} is used in the \texttt{SELECT} clause.
  \item The queries about the cost of or drug routes use \texttt{DISTINCT}.
  \item When retrieving a lab value or vital sign, only the value is returned, not the unit of measurement.
  \item \texttt{DENSE\_RANK} is used for ranking questions, meaning multiple items with the same ranks can be returned together. For example, a query asking about the top three frequent diagnoses may retrieve more than three diagnosis names when items with the same rank exist in the answer. Additionally, the retrieved results can be fewer than the expected number \texttt{N} when the number of diagnoses under some conditions is smaller than the expected number.
  \item When a question is related to both death and diagnosis, only the first diagnosis time is considered.
  \item When calculating the \texttt{N}-year survival rate, if a death record exists between the first diagnosis time and \texttt{N} years later, it counts as death. But if there is no death record within \texttt{N} years or the death happens after \texttt{N} years, it counts as survived.
  \item The current time and normal ranges of vital signs are post-processed after SQL generation so that they are independent of the modeling pipeline when the value changes.
  \item The vital signs we consider in the dataset are body temperature, SaO2, heart rate, respiratory rate, and blood pressures (systemic systolic, diastolic, and mean).
\end{itemize}

\subsubsection{Assumptions in MIMIC-III}

\begin{itemize}
  \item Diagnosis and procedure times are not available in the original database. To address this, we manually set diagnosis time as hospital admission time and procedure time as hospital discharge time. Thus, questions asking about current patients' procedure time are excluded in the MIMIC-III questions. 
  \item Among many items in the CHARTEVENTS table, we only use weight, height, and the seven vital sign values.
  \item We use INPUTEVENTS\_CV instead of INPUTEVENTS\_MV for input events as it contains more records and one time column per record (charttime), which is more closely aligned with eICU's intakeoutput table.
  \item For input and output events, we only use values stored in milliliters (mL) in the case of numerical reasoning within or between input and output events.
\end{itemize}

\subsubsection{Assumptions in eICU}

\begin{itemize}
  \item Diagnosis and treatment tables have path-based names for each record. Instead of using them directly, the paths are further pre-processed to have shorter names.
  \item For vital signs, we choose the vitalperiodic table as it contains more records.
  \item Questions about drug dose are not considered in eICU as drug doses are stored in free-text (values and units are mixed).
\end{itemize}

\subsection{Mapping Between Condition Value Slots and Column Names}
\label{sec:schema_mapping}

The mapping between condition value slots and column names in both MIMIC-III and eICU is shown in Table~\ref{tab:schema_mapping}.

\input{supp/schema_mapping}

\newpage
\subsection{Comparison Between \texttt{JOIN}-based and Nesting-based Queries}
\label{sec:join_vs_nested}

\begin{figure}[t!]
\centering
\includegraphics[width=\linewidth]{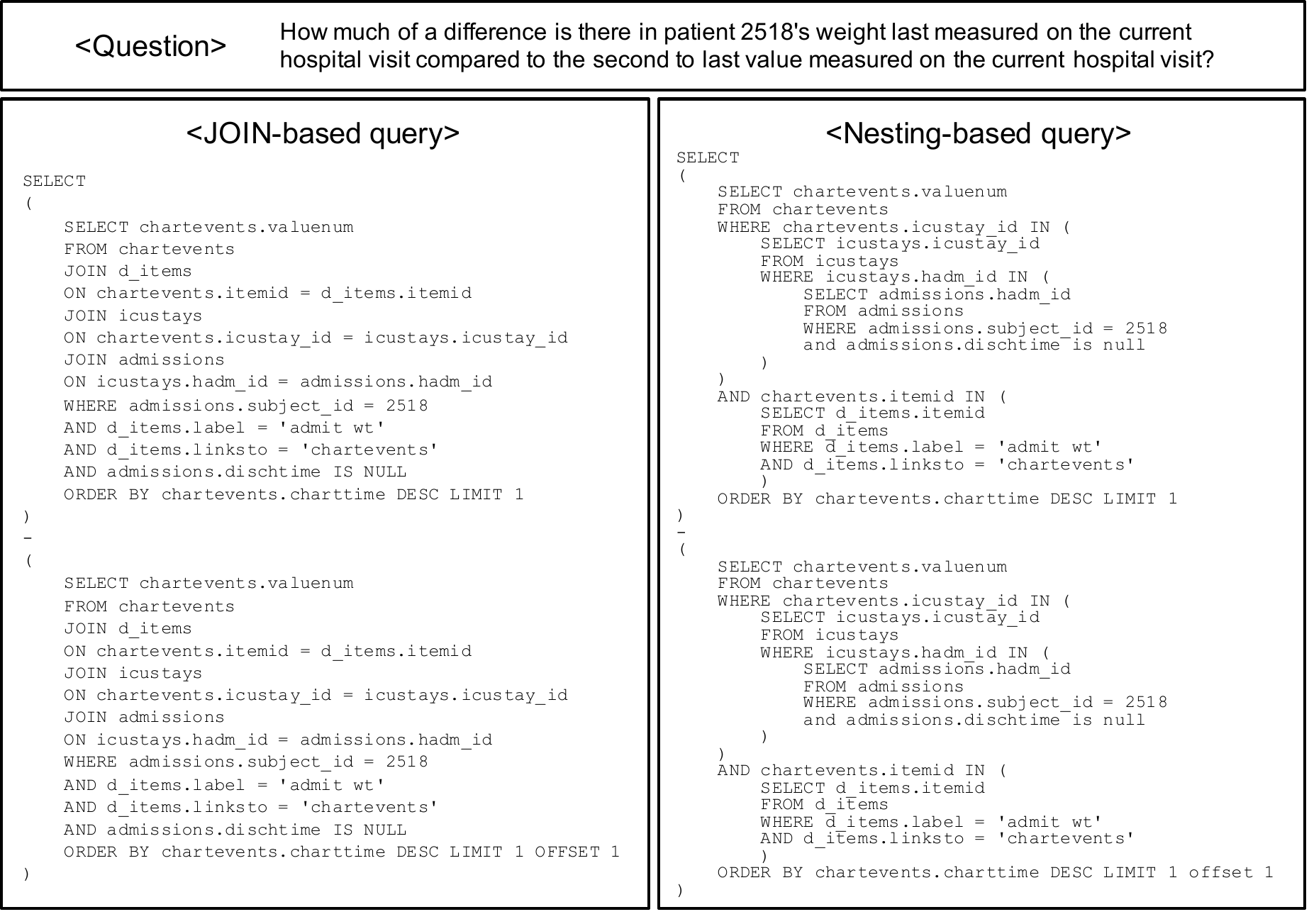}
\caption{JOIN-based and nesting-based queries.}
\label{figure:join_based_and_nested}
\end{figure}

Unlike most other semantic parsing datasets, the SQL queries in EHRSQL are labeled in a nested manner. Table~\ref{figure:join_based_and_nested} shows a comparison between queries with the naive use of \texttt{JOIN} and nesting. In terms of query length, \texttt{JOIN}-based queries are much shorter, but it is hard to follow the semantics in the queries. However, even if the length of the queries is long, T5 models are able to fully generate a long sequence of SQL (see Table~\ref{tab:qualitative_result_complexity} for qualitative results). As for execution speed, the naive use of \texttt{JOIN} takes almost four times slower than a nesting-based query in some cases (0.04 vs. 0.15 secs), as shown in Figure~\ref{figure:join_based_and_nested}.

\section{Database Pre-processing Details}

\subsection{Database Pre-processing Rules}

\begin{itemize}
  \item Patients aged 11 to 89 are included. In eICU, ages recorded as ``> 89'' are mapped to 90 and thus retained.
  \item 1,000 patients are sampled to cover a greater number of medical events (MIMICSQL and emrKBQA use 100 patients).
  \item When the same type of value has multiple units of measurement, only the value with the most common unit is retained and other values are removed from the database.
  \item All records are lower-cased.
\end{itemize}

\subsection{Time-shifting Process}
\label{sec:time_shifting}

To include questions with relative time expressions, we manually time-shift each patient's hospital records. Specifically, we sample a random time point (between 2100 and 2105) to set the time of the first hospital visit. Then, we time-shift the whole patient records to the sampled time point while keeping all the record intervals the same. Additionally, we constrain the number of current patients to 10\% of the total number of patients in patient sampling since any questions can be asked with relative time expressions, such as yesterday and last month.

\subsection{De-identification Process}
\label{sec:deidentification}

Even though MIMIC-III and eICU are de-identified databases, they still contain patient-specific information. In our case, when one or more condition values are sampled along with a patient ID, there is a risk that the question and its paired SQL query might reveal patient-specific information. To avoid such risk, we randomly shuffle records of diagnosis, procedure, lab tests, prescription, chart events, input events, output events, microbiology tests, care units, and ward IDs across patients in the database, while keeping the time of the records the same.

\section{Template Paraphrasing Pipeline}
\label{sec:template_paraphrasing}

\begin{figure}[h!]
\centering
\includegraphics[width=\linewidth]{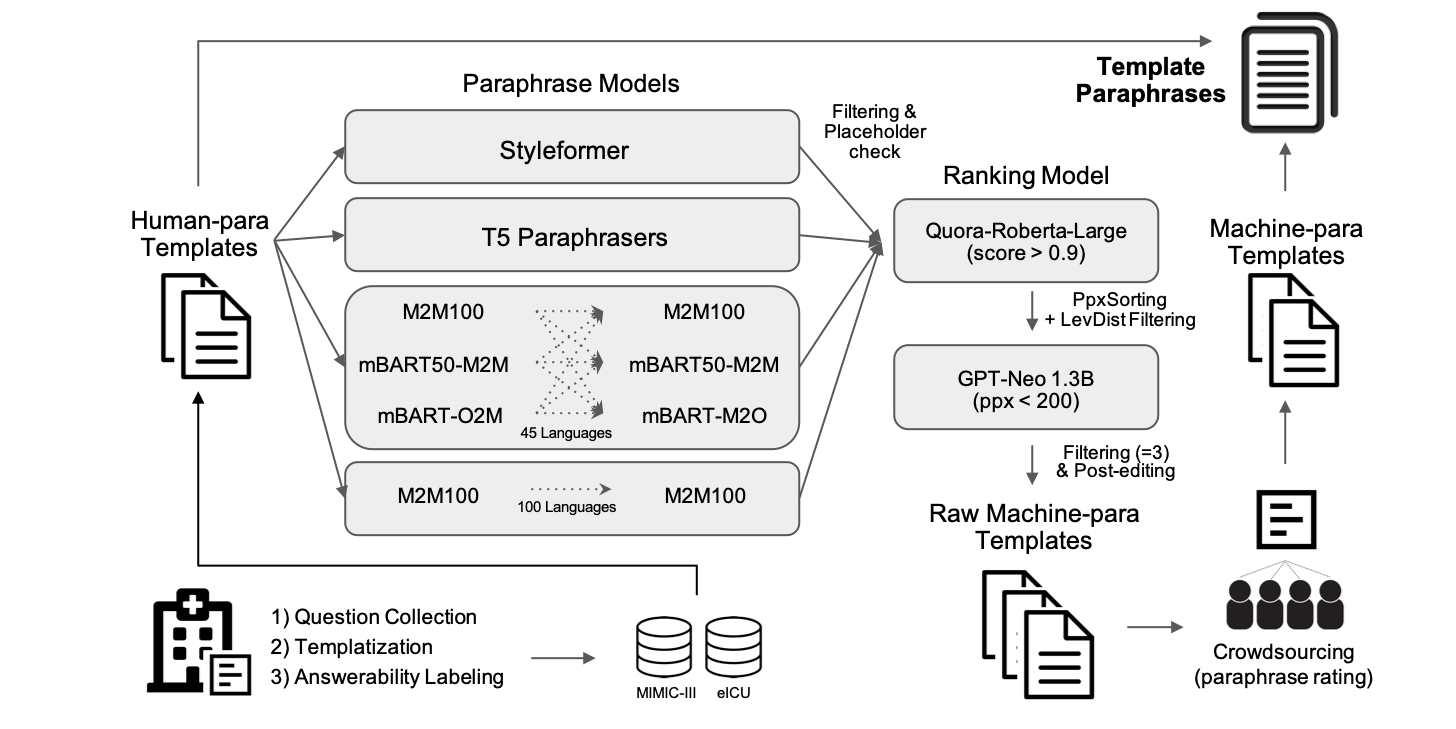}
\caption{Template paraphrasing pipeline.}
\label{figure:paraphrase_generation}
\end{figure}

The overall template paraphrasing pipeline is illustrated in Figure~\ref{figure:paraphrase_generation}. 



  

\section{Data Splitting}
\label{sec:data_splitting}

We constrain the training, validation, and test sets to contain all question templates, but the template paraphrases do not overlap between the splits. 

\begin{itemize}
  \item Training set: We provide more text-to-SQL pairs to question templates with a greater number of slots. Specifically, question templates with fewer than three slots are assigned thirty to sixty pairs. Questions with three or more slots and fewer than five receive forty to eighty pairs; questions with five or more slots receive fifty to one hundred pairs per question template.
  \item Validation set: The number of pairs per question template is sampled between four and five.
  \item Test set: Data sampling rules are identical to those of the validation set, except that we assign higher weights on the question templates that are considered important, which are labeled in ``high,'' ``medium,'' ``low,'' and ``n/a'' (not available) by a physician. With a score of 3 for ``high,'' 2 for ``medium,'' 1 for the rest, the final number of pairs in the test set is multiplied by the importance score for each question template. 
  \item Unanswerable question: They are assigned to the validation and test sets so that each split makes up 33\%.
\end{itemize}

\section{Training Details}
\label{sec:training_detail}

We use pre-trained T5-base models from Hugging Face\footnote{\url{https://huggingface.co/}} for both the no schema and schema versions. Without a schema, the model is trained to translate from natural questions to SQL queries. With a schema, we additionally append schema information to the questions. The training and evaluation configurations are in Table~\ref{tab:hyperparam}. All models are trained on NVIDIA GeForce RTX 3090s.

\begin{table}[h!]
\centering
\caption{T5-base fine-tuning configurations.}
\begin{small}
\begin{tabular}{cc}
\toprule
\multicolumn{2}{c}{Training} \\
\midrule
Total training step & 100,000 \\
Batch size & 32 \\
Max length & 512 \\
Optimizer & Adam \\
Learning rate & 1e-4 \\
Learning rate scheduler & Fixed \\
Max gradient norm & 1.0 \\
Weight decay & 0.1 \\
Validation step & 5,000 \\
\midrule
\multicolumn{2}{c}{Evaluation} \\
\midrule
Num beams & 5 \\
Repetition penalty & 1.0 \\
Length penalty & 1.0 \\
\bottomrule
\end{tabular}
\end{small}
\label{tab:hyperparam}
\end{table} 


\section{Qualitative Results}
\label{sec:qualitative_result}

\subsection{SQL Generation by Question Complexity}

Table~\ref{tab:qualitative_result_complexity} shows samples of generated SQL queries by the number of slots. A T5-base model trained on MIMIC-III can generate a very long sequence of SQL queries if they are seen in the training data. In most cases, the errors do not come from generating complex, long nested queries, but from failures in schema linking (identifying references of columns, tables, and condition values in natural utterances).

\begin{table}[h!]
\centering
\caption{SQL generation results by complexity.}
\begin{small}
\begin{adjustbox}{width=\columnwidth,center}  
\begin{tabular}{cccc}
\toprule
\textbf{Number of slots} & \textbf{Question} &\textbf{Real SQL} & \textbf{Generated SQL} \\
\midrule
1 
    & \makecell[l]{what is the method for ingestion of danazol?} 
    & \makecell[l]{select distinct prescriptions.route from prescriptions \\ where prescriptions.drug = \textquotesingle danazol\textquotesingle}
    & \makecell[l]{select distinct prescriptions.route from prescriptions \\ where prescriptions.drug = \textquotesingle\textcolor{red}{ingestion of} danazol\textquotesingle} \\
\midrule    
2 
    & \makecell[l]{how many patients were given temporary \\ tracheostomy?} 
    & \makecell[l]{select count( distinct admissions.subject\_id ) from \\ admissions where admissions.hadm\_id in ( select \\ procedures\_icd.hadm\_id from procedures\_icd where \\ procedures\_icd.icd9\_code = ( select \\ d\_icd\_procedures.icd9\_code from d\_icd\_procedures where \\ d\_icd\_procedures.short\_title = \textquotesingle temporary tracheostomy\textquotesingle ) )}
    & \makecell[l]{select count( distinct admissions.subject\_id ) from \\ admissions where admissions.hadm\_id in ( select \\ procedures\_icd.hadm\_id from procedures\_icd where \\ procedures\_icd.icd9\_code = ( select \\ d\_icd\_procedures.icd9\_code from d\_icd\_procedures where \\ d\_icd\_procedures.short\_title = \textquotesingle \textcolor{red}{procedures\_icd}\textquotesingle ) )} \\
\midrule    
2 
    & \makecell[l]{tell me the length of the hospital stay of \\ the last hospital stay of patient 3745.} 
    & \makecell[l]{select strftime(\textquotesingle\%J\textquotesingle,admissions.dischtime) - \\ strftime(\textquotesingle\%J\textquotesingle,admissions.admittime) from admissions where \\ admissions.subject\_id = 3745 and admissions.dischtime is \\ 
    not null order by admissions.admittime desc limit 1 }
    & \makecell[l]{select strftime(\textquotesingle\%J\textquotesingle,admissions.dischtime) - \\ strftime(\textquotesingle\%J\textquotesingle,admissions.admittime) from admissions where \\ admissions.subject\_id = 3745 and admissions.dischtime is \\ 
    not null order by admissions.admittime desc limit 1 } \\
\midrule    
4 
    & \makecell[l]{tell me the percentile of mcv 97.0 in \\ patients of the same age as patient 2127 \\ during the last hospital visit?} 
    & \makecell[l]{select distinct t1.c1 from ( select labevents.valuenum, \\ percent\_rank() over ( order by labevents.valuenum ) as c1 \\ from labevents where labevents.itemid in ( select \\ d\_labitems.itemid from d\_labitems where d\_labitems.label \\ =\textquotesingle mcv\textquotesingle ) and labevents.hadm\_id in ( select \\ admissions.hadm\_id from admissions where admissions.age = ( \\ select admissions.age from admissions where \\ admissions.subject\_id = 2127 and admissions.dischtime is \\ not null order by admissions.admittime desc limit 1 ) ) ) \\ as t1 where t1.valuenum = 97.0}
    & \makecell[l]{select distinct t1.c1 from ( select labevents.valuenum, \\ percent\_rank() over ( order by labevents.valuenum ) as c1 \\ from labevents where labevents.itemid in ( select \\ d\_labitems.itemid from d\_labitems where d\_labitems.label \\ =\textquotesingle mcv\textquotesingle ) and labevents.hadm\_id in ( select \\ admissions.hadm\_id from admissions where admissions.age = ( \\ select admissions.age from admissions where \\ admissions.subject\_id = 2127 and admissions.dischtime is \\ not null order by admissions.admittime desc limit 1 ) ) ) \\ as t1 where t1.valuenum = 97.0} \\
\midrule
6 
    & \makecell[l]{what\textquotesingle s the potassium change/difference of \\ patient 99747 second measured on the last \\ hospital visit compared to the first value \\ measured on the last hospital visit?} 
    & \makecell[l]{select ( select labevents.valuenum from labevents where \\ labevents.hadm\_id in ( select admissions.hadm\_id from \\ admissions where admissions.subject\_id = 99747 and \\ admissions.dischtime is not null order by \\ admissions.admittime desc limit 1 ) and labevents.itemid in \\ ( select d\_labitems.itemid from d\_labitems where \\ d\_labitems.label = \textquotesingle potassium\textquotesingle ) order by \\ labevents.charttime asc limit 1 offset 1 ) - ( select \\ labevents.valuenum from labevents where labevents.hadm\_id \\ in ( select admissions.hadm\_id from admissions where \\ admissions.subject\_id = 99747 and admissions.dischtime is \\ not null order by admissions.admittime desc limit 1 ) and \\ labevents.itemid in ( select d\_labitems.itemid from \\ d\_labitems where d\_labitems.label = \textquotesingle potassium\textquotesingle ) order by \\ labevents.charttime asc limit 1 )}
    & \makecell[l]{select ( select labevents.valuenum from labevents where \\ labevents.hadm\_id in ( select admissions.hadm\_id from \\ admissions where admissions.subject\_id = 99747 and \\ admissions.dischtime is not null order by \\ admissions.admittime desc limit 1 ) and labevents.itemid in \\ ( select d\_labitems.itemid from d\_labitems where \\ d\_labitems.label = \textquotesingle potassium\textquotesingle ) order by \\ labevents.charttime asc limit 1 offset 1 ) - ( select \\ labevents.valuenum from labevents where labevents.hadm\_id \\ in ( select admissions.hadm\_id from admissions where \\ admissions.subject\_id = 99747 and admissions.dischtime is \\ not null order by admissions.admittime desc limit 1 ) and \\ labevents.itemid in ( select d\_labitems.itemid from \\ d\_labitems where d\_labitems.label = \textquotesingle potassium\textquotesingle ) order by \\ labevents.charttime asc limit 1 )} \\
\bottomrule
\end{tabular}
\end{adjustbox}
\end{small}
\label{tab:qualitative_result_complexity}
\end{table}

\subsection{SQL Generation with Different Time Expressions}

Table~\ref{tab:qualitative_result_time} shows generated SQL samples with different time templates. The column \textit{Seen vs. Unseen} indicates whether the exact question template (\textit{Q\_tag}) and time template (\textit{T\_tag}) combination is seen during training. Interestingly, the model can correctly generate SQL queries for both seen and unseen combinations.

\begin{table}[h!]
\centering
\caption{SQL generation results with different time expressions.}
\begin{small}
\begin{adjustbox}{width=\columnwidth,center}  
\begin{tabular}{ccccc}
\toprule
\textbf{Question} & \textbf{Real SQL \& Generated SQL} & \textbf{Q\_tag $\times$ T\_tag} & \textbf{\makecell{Seen vs. \\ Unseen}} \\
\midrule
\makecell[l]{the first care unit of patient 46422 \textcolor{blue}{since} \\\textcolor{blue}{2101} is?} 
    & \makecell[l]{select transfers.careunit from transfers where \\ transfers.hadm\_id in ( select admissions.hadm\_id from \\ admissions where admissions.subject\_id = 46422 ) and \\ transfers.careunit is not null \textcolor{blue}{and} \\ \textcolor{blue}{strftime(\textquotesingle \%Y\textquotesingle,transfers.intime) >= \textquotesingle 2101\textquotesingle} order by \\ transfers.intime asc limit 1}
    & \makecell[c]{what was the \texttt{[time\_filter\_exact1]} careunit of \\ patient \texttt{\{patient\_id\}} \texttt{[time\_filter\_global1]}? 
    \\
    $\times$
    \\
    (\textcolor{blue}{\textquotesingle abs-year-since\textquotesingle}, \textquotesingle\textquotesingle, \textquotesingle\textquotesingle, \textquotesingle exact-first\textquotesingle, \textquotesingle\textquotesingle)}
    & seen \\
\midrule
\makecell[l]{what is the first careunit that patient \\ 53089 stayed \textcolor{blue}{on the last hospital} \\ \textcolor{blue}{encounter}?} 
    & \makecell[l]{select transfers.careunit from transfers where \\ transfers.hadm\_id in ( select admissions.hadm\_id from \\ admissions where admissions.subject\_id = 53089 \textcolor{blue}{and} \\ \textcolor{blue}{admissions.dischtime is not null order by} \\ \textcolor{blue}{admissions.admittime desc limit 1} ) and transfers.careunit \\ is not null order by transfers.intime asc limit 1}
    & \makecell[c]{what was the \texttt{[time\_filter\_exact1]} careunit of \\ patient \texttt{\{patient\_id\}} \texttt{[time\_filter\_global1]}? 
    \\
    $\times$
    \\    
    (\textcolor{blue}{\textquotesingle rel-hosp-last\textquotesingle}, \textquotesingle\textquotesingle, \textquotesingle\textquotesingle, \textquotesingle exact-first\textquotesingle, \textquotesingle\textquotesingle)}
    & unseen \\    
\midrule
\makecell[l]{count the number of patients that were \\ prescribed aspirin ec within 2 months \\ after having received a venous cath nec \\ procedure \textcolor{blue}{until 4 years ago}.} 
    & \makecell[l]{select count( distinct t1.subject\_id ) from ( select \\ admissions.subject\_id, procedures\_icd.charttime from \\ procedures\_icd join admissions on procedures\_icd.hadm\_id = \\ admissions.hadm\_id where procedures\_icd.icd9\_code = ( \\ select d\_icd\_procedures.icd9\_code from d\_icd\_procedures \\ where d\_icd\_procedures.short\_title =\textquotesingle venous cath nec\textquotesingle ) \textcolor{blue}{and} \\ \textcolor{blue}{datetime(procedures\_icd.charttime) <= datetime(\textquotesingle2105-12-31} \\ \textcolor{blue}{23:59:00\textquotesingle,\textquotesingle-4 year\textquotesingle)} ) as t1 join ( select \\ admissions.subject\_id, prescriptions.startdate from \\ prescriptions join admissions on prescriptions.hadm\_id = \\ admissions.hadm\_id where prescriptions.drug = \textquotesingle aspirin ec\textquotesingle \\ \textcolor{blue}{and datetime(prescriptions.startdate) <=} \\ \textcolor{blue}{datetime(\textquotesingle2105-12-31 23:59:00\textquotesingle,\textquotesingle-4 year\textquotesingle)} ) as t2 on \\ t1.subject\_id = t2.subject\_id where t1.charttime < \\ t2.startdate and datetime(t2.startdate) between datetime( \\ t1.charttime) and datetime(t1.charttime,\textquotesingle+2 month\textquotesingle)}
    & \makecell[c]{count the number of patients who were \\ prescribed \texttt{\{drug\_name\}} \\ \texttt{[time\_filter\_within]} after having \\ received a \texttt{\{procedure\_name\}} procedure \\ \texttt{[time\_filter\_global1]}. 
    \\
    $\times$
    \\ (\textcolor{blue}{\textquotesingle rel-year-until\textquotesingle}, \textquotesingle\textquotesingle, \textquotesingle within-n\_month\textquotesingle, \textquotesingle\textquotesingle, \textquotesingle\textquotesingle)}
    & seen \\
\midrule
\makecell[l]{how many patients were prescribed \\ levothyroxine sodium within 2 months \textcolor{blue}{since} \\ \textcolor{blue}{2105} after the procedure of cath base \\ invasv ep test.} 
    & \makecell[l]{select count( distinct t1.subject\_id ) from ( select \\ admissions.subject\_id, procedures\_icd.charttime from \\ procedures\_icd join admissions on procedures\_icd.hadm\_id = \\ admissions.hadm\_id where procedures\_icd.icd9\_code = ( \\ select d\_icd\_procedures.icd9\_code from d\_icd\_procedures \\ where d\_icd\_procedures.short\_title = \textquotesingle cath base invasv ep \\ test\textquotesingle ) \textcolor{blue}{and strftime(\textquotesingle \%Y\textquotesingle,procedures\_icd.charttime) >=} \\ \textcolor{blue}{\textquotesingle 2105\textquotesingle} ) as t1 join ( select admissions.subject\_id, \\ prescriptions.startdate from prescriptions join admissions \\ on prescriptions.hadm\_id = admissions.hadm\_id where \\ prescriptions.drug = \textquotesingle levothyroxine sodium\textquotesingle \, \textcolor{blue}{and} \\ \textcolor{blue}{strftime(\textquotesingle\%Y\textquotesingle,prescriptions.startdate) >= \textquotesingle 2105\textquotesingle} ) as t2 on \\ t1.subject\_id = t2.subject\_id where t1.charttime < \\ t2.startdate and datetime(t2.startdate) between datetime( \\ t1.charttime) and datetime(t1.charttime,\textquotesingle+2 month\textquotesingle)}
    & \makecell[c]{count the number of patients who were \\ prescribed \texttt{\{drug\_name\}} \\ \texttt{[time\_filter\_within]} after having \\ received a \texttt{\{procedure\_name\}} procedure \\ \texttt{[time\_filter\_global1]}. 
    \\
    $\times$
    \\ (\textcolor{blue}{\textquotesingle abs-year-since\textquotesingle}, \textquotesingle\textquotesingle, \textquotesingle within-n\_month\textquotesingle, \textquotesingle\textquotesingle, \textquotesingle\textquotesingle)} 
    & unseen \\
\bottomrule
\end{tabular}
\end{adjustbox}
\end{small}
\label{tab:qualitative_result_time}
\end{table}

\subsection{Falsely Executed and Refused Results}

Table~\ref{tab:false_executed_result} and~\ref{tab:refused_result} show samples of falsely executed and refused questions, respectively. Falsely executed results are retrieved results of the model even though the input question is unanswerable (see Table~\ref{tab:false_executed_result}). These errors are fatal mistakes that a healthcare QA system must avoid, as retrieving incorrect information may lead to wrong clinical decisions. 

Table~\ref{tab:refused_result} shows refused results of the model. In some cases, the model might be able to generate the SQL query, but chooses not to execute it due to low confidence.

\begin{table}[h!]
\centering
\caption{Falsely executed results.}
\begin{small}
\begin{adjustbox}{width=\columnwidth,center}  
\begin{tabular}{ccccc}
\toprule
\textbf{Question} & \textbf{Real SQL} & \textbf{Generated SQL} & \textbf{Retrieved Answer} & \textbf{Comment} \\
\midrule
\makecell[l]{what was the duration of the packed cell \\ transfusion procedure for patient 9566?}
    &  nan
    &  \makecell[l]{select strftime(\textquotesingle \%J\textquotesingle,admissions.dischtime) - \\ strftime(\textquotesingle \%J\textquotesingle,procedures\_icd.charttime) from procedures\_icd \\ where procedures\_icd.icd9\_code = ( select \\ d\_icd\_procedures.icd9\_code from d\_icd\_procedures where \\ d\_icd\_procedures.short\_title = 'packed cell transfusion' ) )}
    &  
    & \makecell[c]{Execution \\ error} \\
\midrule
\makecell[l]{how long did it take to get venous cath nec for \\ patient 31854?}
    &  nan
    &  \makecell[l]{select strftime(\textquotesingle \%J\textquotesingle,admissions.dischtime) - \\ strftime(\textquotesingle \%J\textquotesingle,admissions.admittime) from admissions where \\ admissions.subject\_id = 31854 and admissions.dischtime is \\ not null order by admissions.admittime asc limit 1}
    &  [(8.61111111100763,)]
    & \makecell[c]{Incorrect answer \\ retrieved} \\    
\bottomrule
\end{tabular}
\end{adjustbox}
\end{small}
\label{tab:false_executed_result}
\end{table}

\begin{table}[h!]
\centering
\caption{Refused results of both correct and incorrect samples.}
\begin{small}
\begin{adjustbox}{width=\columnwidth,center}  
\begin{tabular}{cccc}
\toprule
\textbf{Question} & \textbf{Real SQL} & \textbf{Generated SQL} & \textbf{Comment} \\
\midrule
\makecell[l]{how many patients were prescribed with \\ magnesium sulfate within the same month after \\ the treatment of cont inv mec ven <96 hrs?}
    & \makecell[l]{select count( distinct t1.subject\_id ) from ( select \\ admissions.subject\_id, procedures\_icd.charttime from \\ procedures\_icd join admissions on procedures\_icd.hadm\_id = \\ admissions.hadm\_id where procedures\_icd.icd9\_code = \\ ( select d\_icd\_procedures.icd9\_code from d\_icd\_procedures \\ where d\_icd\_procedures.short\_title = \textquotesingle cont inv mec ven <96 \\ hrs\textquotesingle ) ) as t1 join ( select admissions.subject\_id, \\ prescriptions.startdate from prescriptions join admissions \\ on prescriptions.hadm\_id = admissions.hadm\_id where \\ prescriptions.drug =\textquotesingle magnesium sulfate\textquotesingle ) as t2 on \\ t1.subject\_id = t2.subject\_id where t1.charttime < \\ t2.startdate and datetime(t1.charttime,'start of month') = \\ datetime(t2.startdate,\textquotesingle start of month\textquotesingle)} 
    & \makecell[l]{select count( distinct t1.subject\_id ) from ( select \\ admissions.subject\_id, procedures\_icd.charttime from \\ procedures\_icd join admissions on procedures\_icd.hadm\_id = \\ admissions.hadm\_id where procedures\_icd.icd9\_code = \\ ( select d\_icd\_procedures.icd9\_code from d\_icd\_procedures \\ where d\_icd\_procedures.short\_title = \textquotesingle cont inv mec ven <96 \\ hrs\textquotesingle ) ) as t1 join ( select admissions.subject\_id, \\ prescriptions.startdate from prescriptions join admissions \\ on prescriptions.hadm\_id = admissions.hadm\_id where \\ prescriptions.drug =\textquotesingle magnesium sulfate\textquotesingle ) as t2 on \\ t1.subject\_id = t2.subject\_id where t1.charttime < \\ t2.startdate and datetime(t1.charttime,'start of month') = \\ datetime(t2.startdate,\textquotesingle start of month\textquotesingle)} 
    & \makecell[c]{Correctly generated \\ but refused} \\
\midrule
\makecell[l]{how is naproxen taken? }
    & \makecell[l]{select distinct prescriptions.route from prescriptions \\ where prescriptions.drug = \textquotesingle naproxen\textquotesingle} 
    & \makecell[l]{select inputevents\_cv.amount from inputevents\_cv where \\ inputevents\_cv.icustay\_id in ( select icustays.icustay\_id \\ from icustays where icustays.hadm\_id in ( select \\ admissions.hadm\_id from admissions where \\ admissions.subject\_id = naproxen ) ) and \\ inputevents\_cv.itemid in ( select d\_items.itemid from \\ d\_items where d\_items.label = \textquotesingle niroxen\textquotesingle and d\_items.linksto \\ = \textquotesingle inputevents\_cv\textquotesingle )} 
    & \makecell[c]{Incorrectly generated \\ and refused} \\
\bottomrule
\end{tabular}
\end{adjustbox}
\end{small}
\label{tab:refused_result}
\end{table}

\newpage
\subsection{Entropy Distribution of the Model Outcome}
\label{sec:result_entropy}

\begin{figure}[h!]
\centering
\includegraphics[scale=0.5]{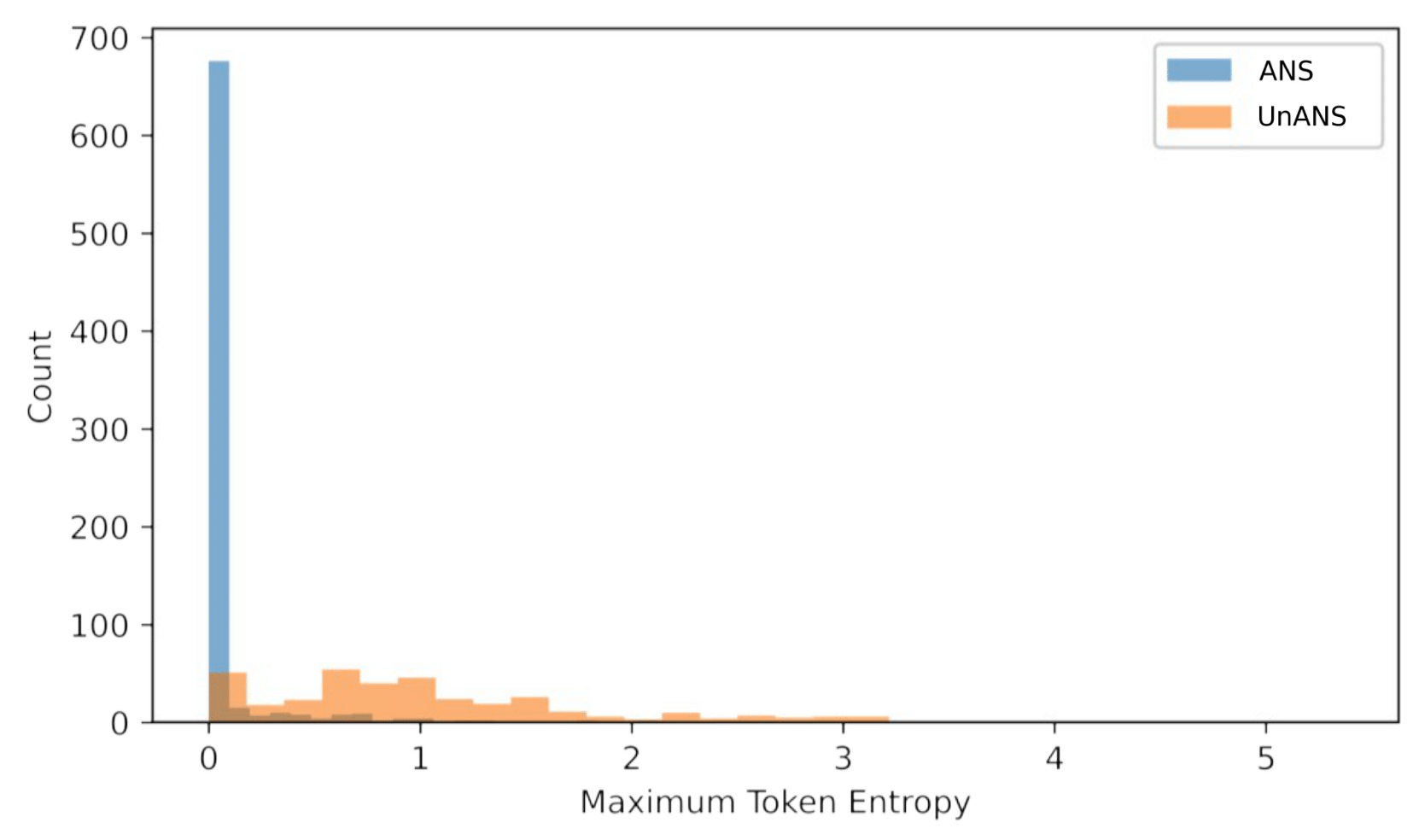}
\caption{Distribution of entropy values generated from T5.}
\label{figure:max_tok_png}
\end{figure}

Figure~\ref{figure:max_tok_png} shows the maximum entropy values generated from T5. As the ground-truth labels (ANS: answerable; UnANS: unanswerable) indicate, the distributions of entropy values between answerable and unanswerable questions are significantly different (<\;0.001 with the Mann-Whitney U test). Similar patterns are observed in models trained on eICU.



\section{Author statement}

The authors of this paper bear all responsibility in case of violation of rights, etc. associated with the EHRSQL dataset.

%% file: supp/datasheet_for_dataset.tex
\subsection{Motivation}

\begin{itemize}
  \item  For what purpose was the dataset created?
  \\ EHRSQL is created to serve as a benchmark for trustworthy question answering systems on structured data in electronic health records (EHRs).
  \item Who created the dataset (e.g., which team, research group) and on
behalf of which entity (e.g., company, institution, organization)?
  \\ The authors of this paper.
  \item Who funded the creation of the dataset? If there is an associated
grant, please provide the name of the grantor and the grant name and
number.
  \\ This work was supported by Institute of Information \& Communications Technology Planning \& Evaluation (IITP) grant (No.2019-0-00075, Artificial Intelligence Graduate School Program(KAIST)), National Research Foundation of Korea (NRF) grant (NRF-2020H1D3A2A03100945) and Data Voucher grant (2021-DV-I-P-00114), funded by the Korea government (MSIT).
\end{itemize}

\subsection{Composition}

\begin{itemize}
  \item  What do the instances that comprise the dataset represent (e.g., documents, photos, people, countries)?
  \\ EHRSQL contains natural questions and their corresponding SQL queries (text).
  \item How many instances are there in total (of each type, if appropriate)?
  \\ There are about 24.4K instances (22.5K answerable; 1.9K unanswerable). 
  \item Does the dataset contain all possible instances or is it a sample (not necessarily random) of instances from a larger set?
  \\ We conducted a poll at a university hospital and collected a wide range of questions frequently asked on the structured EHR data. To reflect as many questions as possible, we templatized them and ensured that the final dataset contained all the question templates we created.
  \item  What data does each instance consist of?
  \\ The dataset contains question-SQL pairs if the question is answerable. Unanswerable questions do not have SQL labels.
  \item  Is there a label or target associated with each instance?
  \\ Labels are SQL queries.
  \item  Is any information missing from individual instances? If so, please provide a description, explaining why this information is missing (e.g., because it was unavailable). This does not include intentionally removed information, but might include, e.g., redacted text.
  \\ N/A.
  \item  Are relationships between individual instances made explicit (e.g., users’ movie ratings, social network links)?
  \\ N/A.
  \item  Are there recommended data splits (e.g., training, development/validation, testing)?
  \\ See Section~\ref{sec:data_splitting}.
  \item  Are there any errors, sources of noise, or redundancies in the dataset?
  \\ Question templates are created to have slots that are later filled with pre-defined values and records from the database. As a result, final questions can sound unnatural or be grammatically incorrect depending on the sampled values (\textit{e.g.}, verb tense, articles, etc.).
  \item  Is the dataset self-contained, or does it link to or otherwise rely on external resources (e.g., websites, tweets, other datasets)?
  \\ The labeled SQL queries rely on two open source databases: MIMIC-III (version 1.4)\footnote{\url{https://physionet.org/content/mimiciii/1.4/}} and eICU (version 2.0)\footnote{\url{https://physionet.org/content/eicu-crd/2.0/}}, which are accessible on PhysioNet\footnote{\url{https://physionet.org/}}.
  \item  Does the dataset contain data that might be considered confidential (e.g., data that is protected by legal privilege or by doctor– patient confidentiality, data that includes the content of individuals’ non-public communications)?
  \\ N/A.
  \item  Does the dataset contain data that, if viewed directly, might be offensive, insulting, threatening, or might otherwise cause anxiety?
  \\ N/A.
  \item Does the dataset relate to people?
  \\ Yes.
  \item  Does the dataset identify any subpopulations (e.g., by age, gender)? \\
  EHRSQL is based on patients in MIMIC-III and eICU. MIMIC-III includes over forty thousand patients who stayed in critical care units of the Beth Israel Deaconess Medical Center between 2001 and 2012. eICU contains patients who were discharged between 2014 and 2015 in multiple critical care units in the United States.
  \item  Is it possible to identify individuals (i.e., one or more natural persons), either directly or indirectly (i.e., in combination with other data) from the dataset?
  \\ Even though MIMIC-III and eICU are already de-identified datasets, we further corrupted patient-specific information to avoid any chance of recovering a patient's identity. See Section~\ref{sec:deidentification} for details.
  \item  Does the dataset contain data that might be considered sensitive in any way (e.g., data that reveals race or ethnic origins, sexual orientations, religious beliefs, political opinions or union memberships, or locations; financial or health data; biometric or genetic data; forms of government identification, such as social security numbers; criminal history)?
  \\ The dataset is already de-identified.  
\end{itemize}

\subsection{Collection Process}

\begin{itemize}
  \item  How was the data associated with each instance acquired?
  \\ We collaborated with the Konyang University Hospital\footnote{\url{https://www.kyuh.ac.kr/eng/}} and conducted a poll to collect real-world questions that are frequently asked on the structured EHR data.
  \item  What mechanisms or procedures were used to collect the data (e.g., hardware apparatuses or sensors, manual human curation, software programs, software APIs)?
  \\ We used the website SurveyMonkey\footnote{\url{www.surveymonkey.com}} to create a poll and collect the responses. After the poll, we used Excel, Google Sheets, and Python to process and label the collected data.
  \item  If the dataset is a sample from a larger set, what was the sampling strategy (e.g., deterministic, probabilistic with specific sampling probabilities)?
  \\ When it involves sampling (\textit{e.g.,} data splitting and patient de-identification), we sampled with a fixed random seed.
  \item   Who was involved in the data collection process (e.g., students, crowdworkers, contractors) and how were they compensated (e.g., how much were crowdworkers paid)?
  \\ There were three parts that required human involvement in the data collection process: poll for the question collection, SQL labeling, and quality checking for machine-paraphrased text. For the poll, we provided a \$10 worth of coffee gift card to all poll respondents. For SQL labeling, the authors in the paper manually labeled SQL queries based on the database schemas. We did not hire any crowd worker for this task because the databases contain patient-specific information, and the SQL labeling process required numerous assumptions (\textit{e.g.}, choice of SQL operations, schema linking, etc.). Lastly, we hired crowd workers to check the quality of machine paraphrased text, who were paid approximately \$18 per hour.
  \item  Over what timeframe was the data collected?
  \\ The poll was conducted in February of 2021, but the results do not depend much on the date of data collection.
  \item  Were any ethical review processes conducted (e.g., by an institutional review board)?
  \\ N/A.
  \item Does the dataset relate to people?
  \\ Yes.  
  \item  Did you collect the data from the individuals in question directly, or obtain it via third parties or other sources (e.g., websites)?
  \\ We directly collected the data through a poll.
  \item  Were the individuals in question notified about the data collection?
  \\ Yes. The poll respondents were notified about the use of data. The actual website we used for the poll is here\footnote{\url{https://www.surveymonkey.com/r/Preview/?sm=hv3JkWYLdzXq2G8m_2Bh8yXI8Q_2FHVOzmZHcwFs7D5WhDYPQwgBHaa7OZXASgLWXsBw}}. The poll was conducted in Korean.
  \item  Did the individuals in question consent to the collection and use of their data?
  \\ The purpose of the poll was announced to hospital staff, and only the staff who were interested in the poll participated.
  \item  If consent was obtained, were the consenting individuals provided with a mechanism to revoke their consent in the future or for certain uses?
  \\ N/A.
  \item  Has an analysis of the potential impact of the dataset and its use on data subjects (e.g., a data protection impact analysis) been conducted?
  \\ The dataset does not have individual-specific information.
\end{itemize}

\subsection{Preprocessing/cleaning/labeling}

\begin{itemize}
  \item  Was any preprocessing/cleaning/labeling of the data done (e.g., discretization or bucketing, tokenization, part-of-speech tagging, SIFT feature extraction, removal of instances, processing of missing values)?
  \\ N/A.
  \item  Was the “raw” data saved in addition to the preprocessed/cleaned/labeled data (e.g., to support unanticipated future uses)?
  \\ N/A.  
  \item  Is the software that was used to preprocess/clean/label the data available?
  \\ Preprocessing, cleaning, and labeling are done via Excel, Google Sheets, and Python.
\end{itemize}

\subsection{Uses}

\begin{itemize}
  \item  Has the dataset been used for any tasks already?
  \\ No.
  \item  Is there a repository that links to any or all papers or systems that use the dataset?
  \\ No.
  \item  What (other) tasks could the dataset be used for?
  \\ In addition to solving trustworthy semantic parsing, the seed questions themselves can be a good starting point for any healthcare table-based question answering tasks.
  \item  Is there anything about the composition of the dataset or the way it was collected and preprocessed/cleaned/labeled that might impact future uses?
  \\ N/A.
  \item  Are there tasks for which the dataset should not be used?
  \\ N/A.
\end{itemize}

\subsection{Distribution}

\begin{itemize}
  \item  Will the dataset be distributed to third parties outside of the entity (e.g., company, institution, organization) on behalf of which the dataset was created?
  \\ No.
  \item  How will the dataset be distributed (e.g., tarball on website, API, GitHub)?
  \\ The dataset is released at~\url{https://github.com/glee4810/EHRSQL}.
  \item  When will the dataset be distributed?
  \\ Now.
  \item  Will the dataset be distributed under a copyright or other intellectual property (IP) license, and/or under applicable terms of use (ToU)?
  \\ The dataset is released under CC BY 4.0 license.
  \item Have any third parties imposed IP-based or other restrictions on the data associated with the instances?
  \\ No.
  \item Do any export controls or other regulatory restrictions apply to the dataset or to individual instances?
  \\ No.
\end{itemize}

\subsection{Maintenance}

\begin{itemize}
  \item  Who will be supporting/hosting/maintaining the dataset?
  \\ The authors of this paper.
  \item  How can the owner/curator/manager of the dataset be contacted (e.g., email address)?
  \\ Contact the first author (\url{gyubok.lee@kaist.ac.kr}) or other authors.
  \item  Is there an erratum?
  \\ No.
  \item  Will the dataset be updated (e.g., to correct labeling errors, add new instances, delete instances)?
  \\ If any correction is needed, we plan to upload a new version.
  \item If the dataset relates to people, are there applicable limits on the retention of the data associated with the instances (e.g., were the individuals in question told that their data would be retained for a fixed period of time and then deleted)? 
  \\ N/A
  \item Will older versions of the dataset continue to be supported/hosted/maintained?
  \\ We plan to maintain the newest version only.
  \item If others want to extend/augment/build on/contribute to the dataset, is there a mechanism for them to do so?
  \\ Contact the authors of the paper.
\end{itemize}

%% file: supp/question_template.tex
\begin{longtblr}
[
  caption = {Full list of answerable question templates.},
  label = {tab:question_template_full},
]
{
  colspec = {X[c,m]X[8,l,m]X[1.2,c,m]},
  colsep = 0.5pt,
  rowhead = 1,
  hlines,
  rows={font=\scriptsize},
  row{1} = {c},
} 
\textbf{Patient scope} & \textbf{Question template} & \textbf{Assumption} \\
None 
    & What is the intake method of \texttt{\{drug\_name\}}? 
    & \\
None 
    & What is the cost of a procedure named \texttt{\{procedure\_name\}}? 
    & \\    
None 
    & What is the cost of a \texttt{\{lab\_name\}} lab test? 
    & \\
None 
    & What is the cost of a drug named \texttt{\{drug\_name\}}? 
    & \\
None 
    & What is the cost of diagnosing \texttt{\{diagnosis\_name\}}? 
    & \\
None 
    & What does \texttt{\{abbreviation\}} stand for? 
    & \\
Single 
    & What is the gender of patient \texttt{\{patient\_id\}}? 
    & \\
Single 
    & What is the date of birth of patient \texttt{\{patient\_id\}}? 
    & \\
Single  
    & What was the \texttt{[time\_filter\_exact1]} length of hospital stay of patient \texttt{\{patient\_id\}}? 
    & Only current patients \\
Single  
    & What is the change in the weight of patient \texttt{\{patient\_id\}} from the \texttt{[time\_filter\_exact2]} value measured  \texttt{[time\_filter\_global1]}? 
    & \\ 
Single
    & What is the change in the weight of patient \texttt{\{patient\_id\}} from the \texttt{[time\_filter\_exact2]} value measured \texttt{[time\_filter\_global2]} compared to the \texttt{[time\_filter\_exact1]} value measured \texttt{[time\_filter\_global1]}? 
    & \\
Single 
    & What is the change in the value of \texttt{\{lab\_name\}} of patient \texttt{\{patient\_id\}} from the \texttt{[time\_filter\_exact2]} value  measured \texttt{[time\_filter\_global2]} compared to the \texttt{[time\_filter\_exact1]} value measured \texttt{[time\_filter\_global1]}?
    & \\
Single
    & What is the change in the \texttt{\{vital\_name\}} of patient \texttt{\{patient\_id\}} from the \texttt{[time\_filter\_exact2]} value measured \texttt{[time\_filter\_global2]} compared to the \texttt{[time\_filter\_exact1]} value measured \texttt{[time\_filter\_global1]}? 
    & \\
Single
    & Is the value of \texttt{\{lab\_name\}} of patient \texttt{\{patient\_id\}} \texttt{[time\_filter\_exact2]} measured \texttt{[time\_filter\_global2]} \texttt{[comparison]} than the \texttt{[time\_filter\_exact1]} value measured \texttt{[time\_filter\_global1]}? 
    & \\
Single
    & Is the \texttt{\{vital\_name\}} of patient \texttt{\{patient\_id\}} \texttt{[time\_filter\_exact2]} measured \texttt{[time\_filter\_global2]} \texttt{[comparison]} than the \texttt{[time\_filter\_exact1]} value measured \texttt{[time\_filter\_global1]}?
    & \\
Single 
    & What is\_verb the age of patient \texttt{\{patient\_id\}} \texttt{[time\_filter\_global1]}? 
    & \\
Single 
    & What is\_verb the name of insurance of \texttt{\{patient\_id\}} \texttt{[time\_filter\_global1]}? 
    & \\    
Single 
    & What is\_verb the marital status of patient \texttt{\{patient\_id\}} \texttt{[time\_filter\_global1]}? 
    & \\    
Single 
    & What  percentile is the value of \texttt{\{lab\_value\}} in a \texttt{\{lab\_name\}} lab test among patients of the same age as patient  \texttt{\{patient\_id\}} \texttt{[time\_filter\_global1]}? 
    & \\  
Single 
    & How many \texttt{[unit\_count]} have passed since patient \texttt{\{patient\_id\}} was admitted to the hospital currently? 
    & Only current patient \\
Single
    & How many \texttt{[unit\_count]} have passed since patient \texttt{\{patient\_id\}} was admitted to the ICU currently? 
    & Only current ICU patient \\
Single 
    & How many \texttt{[unit\_count]} have passed since the \texttt{[time\_filter\_exact1]} time patient \texttt{\{patient\_id\}} stayed in careunit \texttt{\{careunit\}} on the current hospital visit?
    & Only current patient \\
Single
    & How many \texttt{[unit\_count]} have passed since the \texttt{[time\_filter\_exact1]} time patient \texttt{\{patient\_id\}} stayed in ward \texttt{\{ward\_id\}} on the current hospital visit? 
    & Only current patient \\
Single 
    & How many \texttt{[unit\_count]} have passed since the \texttt{[time\_filter\_exact1]} time patient \texttt{\{patient\_id\}} received a procedure on the current hospital visit?  
    & Only current patient \\
Single 
    & How many \texttt{[unit\_count]} have passed since the \texttt{[time\_filter\_exact1]} time patient \texttt{\{patient\_id\}} received a \texttt{\{procedure\_name\}} procedure on the current hospital visit? 
    & Only current patient \\
Single 
    & How many \texttt{[unit\_count]} have passed since the \texttt{[time\_filter\_exact1]} time patient \texttt{\{patient\_id\}} was diagnosed with \texttt{\{diagnosis\_name\}} on the current hospital visit? 
    & Only current patient \\
Single
    & How many \texttt{[unit\_count]} have passed since the \texttt{[time\_filter\_exact1]} time patient \texttt{\{patient\_id\}} was prescribed \texttt{\{drug\_name\}} on the current hospital visit? 
    & Only current patient \\
Single
    & How many \texttt{[unit\_count]} have passed since the \texttt{[time\_filter\_exact1]} time patient \texttt{\{patient\_id\}} received a \texttt{\{lab\_name\}} lab test on the current hospital visit? 
    & Only current patient \\
Single 
    & How many \texttt{[unit\_count]} have passed since the \texttt{[time\_filter\_exact1]} time patient \texttt{\{patient\_id\}} had a \texttt{\{intake\_name\}} intake on the current ICU visit? 
    & Only current ICU patient \\
Single
    & What was the \texttt{[time\_filter\_exact1]} hospital admission type of patient \texttt{\{patient\_id\}} \texttt{[time\_filter\_global1]}?
    & \\
Single
    & What was the \texttt{[time\_filter\_exact1]} ward of patient \texttt{\{patient\_id\}} \texttt{[time\_filter\_global1]}? 
    & \\
Single 
    & What was the \texttt{[time\_filter\_exact1]} careunit of patient \texttt{\{patient\_id\}} \texttt{[time\_filter\_global1]}?
    & \\
Single 
    & What was the \texttt{[time\_filter\_exact1]} measured height of patient \texttt{\{patient\_id\}} \texttt{[time\_filter\_global1]}? 
    & \\
Single
    & What was the \texttt{[time\_filter\_exact1]} measured weight of patient \texttt{\{patient\_id\}} \texttt{[time\_filter\_global1]}? 
    & \\
Single 
    & What was the name of the diagnosis that patient \texttt{\{patient\_id\}} \texttt{[time\_filter\_exact1]} received \texttt{[time\_filter\_global1]}?
    & \\
Single 
    & What was the name of the procedure that patient \texttt{\{patient\_id\}} \texttt{[time\_filter\_exact1]} received \texttt{[time\_filter\_global1]}?
    & \\
Single
    & What was the name of the drug that patient \texttt{\{patient\_id\}} was \texttt{[time\_filter\_exact1]} prescribed via \texttt{\{drug\_route\}} route \texttt{[time\_filter\_global1]}?
    & \\
Single 
    & What was the name of the drug that patient \texttt{\{patient\_id\}} was \texttt{[time\_filter\_exact1]} prescribed \texttt{[time\_filter\_global1]}?
    & \\
Single
    & What was the name of the drug that patient \texttt{\{patient\_id\}} was prescribed  \texttt{[time\_filter\_within]} after having been diagnosed with \texttt{\{diagnosis\_name\}} \texttt{[time\_filter\_global1]}? 
    & \\
Single 
    & What was the name of the drug that patient \texttt{\{patient\_id\}} was prescribed  \texttt{[time\_filter\_within]} after having received a \texttt{\{procedure\_name\}} procedure \texttt{[time\_filter\_global1]}? 
    & \\
Single
    & What was the dose of \texttt{\{drug\_name\}} that patient \texttt{\{patient\_id\}} was \texttt{[time\_filter\_exact1]} prescribed \texttt{[time\_filter\_global1]}? 
    & \\
Single 
    & What was the total amount of dose of \texttt{\{drug\_name\}} that patient \texttt{\{patient\_id\}} were prescribed \texttt{[time\_filter\_global1]}?
    & \\
Single 
    & What was the name of the drug that patient \texttt{\{patient\_id\}} were prescribed \texttt{[n\_times]} \texttt{[time\_filter\_global1]}?  
    & \\
Single 
    & What is the new prescription of patient \texttt{\{patient\_id\}} \texttt{[time\_filter\_global2]} compared to the prescription \texttt{[time\_filter\_global1]}? 
    & \texttt{global} filters do not overlap \\ 
Single 
    & What was the \texttt{[time\_filter\_exact1]} measured value of a \texttt{\{lab\_name\}} lab test of patient \texttt{\{patient\_id\}}\texttt{[time\_filter\_global1]}? 
    & \\ 
Single
    & What was the name of the lab test that patient \texttt{\{patient\_id\}} \texttt{[time\_filter\_exact1]} received \texttt{[time\_filter\_global1]}? 
    & \\
Single 
    & what was the \texttt{[agg\_function]} \texttt{\{lab\_name\}} value of patient \texttt{\{patient\_id\}} \texttt{[time\_filter\_global1]}? 
    & \\
Single 
    & What was the name of the allergy that patient \texttt{\{patient\_id\}} had \texttt{[time\_filter\_global1]}?  
    & \\
Single 
    & What was the name of the substance that patient \texttt{\{patient\_id\}} was allergic to \texttt{[time\_filter\_global1]}? 
    & \\
Single 
    & What was the organism name found in the \texttt{[time\_filter\_exact1]} \texttt{\{culture\_name\}} microbiology test of patient \texttt{\{patient\_id\}} \texttt{[time\_filter\_global1]}? 
    & \\ 
Single 
    & What was the name of the specimen that patient \texttt{\{patient\_id\}} was \texttt{[time\_filter\_exact1]} tested \texttt{[time\_filter\_global1]}? 
    & \\ 
Single 
    & What was the name of the intake that patient \texttt{\{patient\_id\}} \texttt{[time\_filter\_exact1]} had \texttt{[time\_filter\_global1]}?
    & \\
Single 
    & What was the total volume of \texttt{\{intake\_name\}} intake that patient \texttt{\{patient\_id\}} received \texttt{[time\_filter\_global1]}? 
    & \\ 
Single
    & What was the total volume of intake that patient \texttt{\{patient\_id\}} received \texttt{[time\_filter\_global1]}? 
    & \\
Single
    & What was the name of the output that patient \texttt{\{patient\_id\}} \texttt{[time\_filter\_exact1]} had \texttt{[time\_filter\_global1]}? 
    & \\ 
Single 
    & What was the total volume of \texttt{\{output\_name\}} output that patient \texttt{\{patient\_id\}} had \texttt{[time\_filter\_global1]}? 
    & \\
Single
    & What was the total volume of output that patient \texttt{\{patient\_id\}} had \texttt{[time\_filter\_global1]}? 
    & \\
Single
    & What is the difference between the total volume of intake and output of patient \texttt{\{patient\_id\}} \texttt{[time\_filter\_global1]}?  
    & \\ 
Single 
    & What was the \texttt{[time\_filter\_exact1]} measured \texttt{\{vital\_name\}} of patient \texttt{\{patient\_id\}\texttt{[time\_filter\_global1]}?} 
    & \\ 
Single
    & What was the \texttt{\{agg\_function\}} \texttt{\{vital\_name\}} of patient \texttt{\{patient\_id\}\texttt{[time\_filter\_global1]}?} 
    & \\ 
Single 
    & What is\_verb the total hospital cost of patient \texttt{\{patient\_id\}} \texttt{[time\_filter\_global1]}? 
    & \\ 
Single
    & When was the \texttt{[time\_filter\_extract1]} hospital admission time of patient \texttt{\{patient\_id\}} \texttt{[time\_filter\_global1]}?   
    & \\ 
Single 
    & When was the \texttt{[time\_filter\_extract1]} hospital admission time that patient \texttt{\{patient\_id\}} was admitted via \texttt{\{admission\_route\}}\texttt{[time\_filter\_global1]}?  
    & \\ 
Single
    & When was the \texttt{[time\_filter\_extract1]} hospital discharge time of patient \texttt{\{patient\_id\}\texttt{[time\_filter\_global1]}?}  
    & \\
Single
    & When was the \texttt{[time\_filter\_extract1]} length of ICU stay of patient \texttt{\{patient\_id\}}? 
    & No current ICU patient \\
Single
    & When was the \texttt{[time\_filter\_extract1]} time that patient \texttt{\{patient\_id\}} was diagnosed with \texttt{\{diagnosis\_name\}} \texttt{[time\_filter\_global1]}?  
    & \\
Single
    & When was the \texttt{[time\_filter\_extract1]} procedure time of patient \texttt{\{patient\_id\}} \texttt{[time\_filter\_global1]}?  
    & \\
Single 
    & When was the \texttt{[time\_filter\_extract1]} time that patient \texttt{\{patient\_id\}} received a \texttt{\{procedure\_name\}} procedure \texttt{[time\_filter\_global1]}?  
    & \\
Single 
    & When was the \texttt{[time\_filter\_extract1]} prescription time of patient \texttt{\{patient\_id\}} \texttt{[time\_filter\_global1]}?  
    & \\
Single 
    & When was the \texttt{[time\_filter\_extract1]} time that patient \texttt{\{patient\_id\}} was prescribed \texttt{\{drug\_name\}} \texttt{[time\_filter\_global1]}?  
    & \\
Single 
    & When was the \texttt{[time\_filter\_extract1]} time that patient \texttt{\{patient\_id\}} was prescribed \texttt{\{drug\_name1\}} and \texttt{\{drug\_name2\}} \texttt{[time\_filter\_within]} \texttt{[time\_filter\_global1]}?  
    & \\
Single
    & When was the \texttt{[time\_filter\_extract1]} time that patient \texttt{\{patient\_id\}} was prescribed a medication via \texttt{\{drug\_route\}} route \texttt{[time\_filter\_global1]}?  
    & \\
Single 
    & When was the \texttt{[time\_filter\_extract1]} time that patient \texttt{\{patient\_id\}} was prescribed a medication via \texttt{\{drug\_route\}} route \texttt{[time\_filter\_global1]}?  
    & \\
Single 
    & When was the \texttt{[time\_filter\_extract1]} lab test of patient \texttt{\{patient\_id\}} \texttt{[time\_filter\_global1]}?  
    & \\
Single 
    & When was the \texttt{[time\_filter\_extract1]} time that patient \texttt{\{patient\_id\}} received a \texttt{\{lab\_test\}} lab test \texttt{[time\_filter\_global1]}?  
    & \\
Single 
    & When was the \texttt{[time\_filter\_extract1]} time that patient \texttt{\{patient\_id\}} had the \texttt{[sort]} value of \texttt{\{lab\_name\}} \texttt{[time\_filter\_global1]}?  
    & \\
Single 
    & When was the \texttt{[time\_filter\_extract1]} microbiology test of patient \texttt{\{patient\_id\}} \texttt{[time\_filter\_global1]}?  
    & \\
Single 
    & When was patient \texttt{\{patient\_id\}}'s \texttt{[time\_filter\_extract1]} \texttt{\{culture\_name\}} microbiology test \texttt{[time\_filter\_global1]}?  
    & \\
Single
    & When was the \texttt{[time\_filter\_extract1]} time that patient \texttt{\{patient\_id\}} had a \texttt{\{intake\_name\}} intake  \texttt{[time\_filter\_global1]}?  
    & \\
Single 
    & When was the \texttt{[time\_filter\_extract1]} intake time of patient \texttt{\{patient\_id\}} \texttt{[time\_filter\_global1]}? 
    & \\
Single
    & When was the \texttt{[time\_filter\_extract1]} time that patient \texttt{\{patient\_id\}} had a \texttt{\{output\_name\}} output \texttt{[time\_filter\_global1]}?  
    & \\
Single 
    & When was the \texttt{[time\_filter\_extract1]} time that patient \texttt{\{patient\_id\}} had a \texttt{\{vital\_name\}} measured \texttt{[time\_filter\_global1]}?  
    & \\
Single 
    & When was the \texttt{[time\_filter\_extract1]} time that the \texttt{\{vital\_name\}} of patient \texttt{\{patient\_id\}} was \texttt{[comparison]} than \texttt{\{vital\_value\}} \texttt{[time\_filter\_global1]}?  
    & \\
Single 
    & When was the \texttt{[time\_filter\_extract1]} time that patient \texttt{\{patient\_id\}} had the \texttt{[sort]} \texttt{\{vital\_name\}} \texttt{[time\_filter\_global1]}?  
    & \\
Single 
    & Has\_verb patient \texttt{\{patient\_id\}} received a \texttt{\{procedure\_name\}} procedure in other than the current hospital \texttt{[time\_filter\_global1]}?  
    & Only sample current patient \\
Single
    & Has\_verb patient \texttt{\{patient\_id\}} been admitted to the hospital \texttt{[time\_filter\_global1]}?
    & \\
Single
    & Has\_verb patient \texttt{\{patient\_id\}} been to an emergency room \texttt{[time\_filter\_global1]}?
    & \\
Single
    & Has\_verb patient \texttt{\{patient\_id\}} received any procedure \texttt{[time\_filter\_global1]}?
    & \\
Single 
    & Has\_verb patient \texttt{\{patient\_id\}} received a \texttt{\{procedure\_name\}} procedure \texttt{[time\_filter\_global1]}?
    & \\
Single 
    & What was the name of the procedure that patient \texttt{\{patient\_id\}} received \texttt{[n\_times]} \texttt{[time\_filter\_global1]}?
    & \\
Single
    & Has\_verb patient \texttt{\{patient\_id\}} received any diagnosis \texttt{[time\_filter\_global1]}?
    & \\
Single 
    & Has\_verb patient \texttt{\{patient\_id\}} been diagnosed with \texttt{\{diagnosis\_name\}} \texttt{[time\_filter\_global1]}?
    & \\
Single 
    & Has\_verb patient \texttt{\{patient\_id\}} been prescribed \texttt{\{drug\_name1\}},\texttt{\{drug\_name2\}}, or \texttt{\{drug\_name3\}} \texttt{[time\_filter\_global1]}?
    & \\
Single 
    & Has\_verb patient \texttt{\{patient\_id\}} been prescribed any medication \texttt{[time\_filter\_global1]}?
    & \\
Single 
    & Has\_verb patient \texttt{\{patient\_id\}} been prescribed \texttt{\{drug\_name\}} \texttt{[time\_filter\_global1]}?
    & \\
Single
    & Has\_verb patient \texttt{\{patient\_id\}} received any lab test \texttt{[time\_filter\_global1]}?
    & \\
Single
    & Has\_verb patient \texttt{\{patient\_id\}} received a \texttt{\{lab\_name\}} lab test \texttt{[time\_filter\_global1]}?
    & \\
Single 
    & Has\_verb patient \texttt{\{patient\_id\}} had any allergy \texttt{[time\_filter\_global1]}?
    & \\
Single
    & Has\_verb patient \texttt{\{patient\_id\}} had any microbiology test result \texttt{[time\_filter\_global1]}?
    & \\
Single
    & Has\_verb patient \texttt{\{patient\_id\}} had any \texttt{\{culture\_name\}} microbiology test result \texttt{[time\_filter\_global1]}?
    & \\
Single
    & Has\_verb there been any organism found in the \texttt{[time\_filter\_extract1]}\texttt{\{culture\_name\}} microbiology test of patient \texttt{\{patient\_id\}} \texttt{[time\_filter\_global1]}?
    & \\
Single
    & Has\_verb patient \texttt{\{patient\_id\}} had any \texttt{\{intake\_name\}} intake \texttt{[time\_filter\_global1]}?
    & \\
Single 
    & Has\_verb patient \texttt{\{patient\_id\}} had any \texttt{\{output\_name\}} output \texttt{[time\_filter\_global1]}?
    & \\
Single
    & Has\_verb the \texttt{\{vital\_name\}} of patient \texttt{\{patient\_id\}} been ever \texttt{[comparison]} than \texttt{\{vital\_value\}} \texttt{[time\_filter\_global1]}?
    & \\
Single
    & Has\_verb the \texttt{\{vital\_name\}} of patient \texttt{\{patient\_id\}} been normal \texttt{[time\_filter\_global1]}?
    & \\
Single
    & List the hospital admission time of patient \texttt{\{patient\_id\}} \texttt{[time\_filter\_global1]}.
    & \\
Single
    & List the \texttt{[unit\_average]} \texttt{[agg\_function]} \texttt{\{lab\_name\}} lab value of patient \texttt{\{patient\_id\}} \texttt{[time\_filter\_global1]}.
    & \\
Single 
    & List the \texttt{[unit\_average]} \texttt{[agg\_function]} weight of patient \texttt{\{patient\_id\}} \texttt{[time\_filter\_global1]}.
    & \\
Single
    & List the \texttt{[unit\_average]} \texttt{[agg\_function]} volume of \texttt{\{intake\_name\}} intake that patient \texttt{\{patient\_id\}} received \texttt{[time\_filter\_global1]}.
    & \\
Single
    & List the \texttt{[unit\_average]} \texttt{[agg\_function]} volume of \texttt{\{output\_name\}} output that patient \texttt{\{patient\_id\}} had \texttt{[time\_filter\_global1]}.
    & \\
Single
    & List the \texttt{[unit\_average]} \texttt{[agg\_function]} \texttt{\{vital\_name\}} of patient \texttt{\{patient\_id\}} \texttt{[time\_filter\_global1]}.
    & \\
Single
    & Count the number of hospital visits of patient \texttt{\{patient\_id\}} \texttt{[time\_filter\_global1]}.
    & \\
Single
    & Count the number of ICU visits of patient \texttt{\{patient\_id\}} \texttt{[time\_filter\_global1]}.
    & \\
Single 
    & Count the number of times that patient \texttt{\{patient\_id\}} received a \texttt{\{procedure\_name\}} procedure \texttt{[time\_filter\_global1]}.
    & \\
Single
    & Count the number of drugs patient \texttt{\{patient\_id\}} was prescribed \texttt{[time\_filter\_global1]}.
    & \\
Single
    & Count the number of times that patient \texttt{\{patient\_id\}} were prescribed \texttt{\{drug\_name\}} \texttt{[time\_filter\_global1]}.
    & \\
Single
    & Count the number of times that patient \texttt{\{patient\_id\}} received a \texttt{\{lab\_name\}} lab test \texttt{[time\_filter\_global1]}.
    & \\
Single
    & Count the number of times that patient \texttt{\{patient\_id\}} had a \texttt{\{intake\_name\}} intake \texttt{[time\_filter\_global1]}.
    & \\
Single
    & Count the number of times that patient \texttt{\{patient\_id\}} had a \texttt{\{output\_name\}} output \texttt{[time\_filter\_global1]}.
    & \\
Group 
    & Count the number of current patients.
    & \\
Group 
    & Count the number of current patients aged \texttt{[age\_group]}.
    & \\
Group 
    & What is the \texttt{[n\_survival\_period]} survival rate of patients diagnosed with \texttt{\{diagnosis\_name\}}?
    & \\
Group 
    & What is the \texttt{[n\_survival\_period]} survival rate of patients who were prescribed \texttt{\{drug\_name\}} after having been diagnosed with \texttt{\{diagnosis\_name\}}?
    & \\
Group 
    & What are the top \texttt{[n\_rank]} diagnoses that have the highest \texttt{[n\_survival\_period]} mortality rate?
    & \\
Group 
    & What is\_verb the \texttt{[agg\_fuction]} total hospital cost that involves a procedure named \texttt{\{procedure\_name\}} \texttt{[time\_filter\_global1]}?
    & \\
Group 
    & What is\_verb the \texttt{[agg\_fuction]} total hospital cost that involves a \texttt{\{lab\_name\}} lab test \texttt{[time\_filter\_global1]}?
    & \\
Group 
    & What is\_verb the \texttt{[agg\_fuction]} total hospital cost that involves a drug named \texttt{\{drug\_name\}} \texttt{[time\_filter\_global1]}?
    & \\
Group 
    & What is\_verb the \texttt{[agg\_fuction]} total hospital cost that involves a diagnosis named \texttt{\{diagnosis\_name\}} \texttt{[time\_filter\_global1]}?
    & \\
Group 
    & List the IDs of patients diagnosed with \texttt{\{diagnosis\_name\}} \texttt{[time\_filter\_global1]}.
    & \\
Group 
    & What is\_verb the \texttt{[agg\_fuction]} \texttt{[unit\_average]}number of patient records diagnosed with \texttt{\{diagnosis\_name\}} \texttt{[time\_filter\_global1]}?
    & \\
Group 
    & Count the number of patients who were dead after having been diagnosed with \texttt{\{diagnosis\_name\}} \texttt{[time\_filter\_within]} \texttt{[time\_filter\_global1]}.
    & \\
Group 
    & Count the number of patients who did not come back to the hospital \texttt{[time\_filter\_within]} after diagnosed with \texttt{\{diagnosis\_name\}} \texttt{[time\_filter\_global1]}.
    & \\
Group 
    & Count the number of patients who were admitted to the hospital \texttt{[time\_filter\_global1]}.
    & \\
Group 
    & Count the number of patients who were discharged from the hospital \texttt{[time\_filter\_global1]}.
    & \\
Group 
    & Count the number of patients who stayed in ward \texttt{\{ward\_id\}} \texttt{[time\_filter\_global1]}.
    & \\
Group 
    & Count the number of patients who stayed in careunit\texttt{\{careunit\}} \texttt{[time\_filter\_global1]}.
    & \\
Group 
    & Count the number of patients who were diagnosed with\texttt{\{diagnosis\_name\}} \texttt{[time\_filter\_within]} after having received a \texttt{\{procedure\_name\}} procedure \texttt{[time\_filter\_global1]}.
    & \\
Group 
    & Count the number of patients who were diagnosed with\texttt{\{diagnosis\_name2\}} \texttt{[time\_filter\_within]} after having been diagnosed with \texttt{\{diagnosed\_name1\}} \texttt{[time\_filter\_global1]}.
    & \\
Group 
    & Count the number of patients who were diagnosed with \texttt{\{diagnosis\_name\}} \texttt{[time\_filter\_global1]}.
    & \\
Group 
    & Count the number of patients who received a \texttt{\{procedure\_name\}} procedure \texttt{[time\_filter\_global1]}.
    & \\
Group 
    & Count the number of patients who received a \texttt{\{procedure\_name\}} procedure \texttt{[n\_times]} \texttt{[time\_filter\_global1]}.
    & \\
Group 
    & Count the number of patients who received a \texttt{\{procedure\_name2\}} procedure \texttt{[time\_filter\_within]} after having received a \texttt{\{procedure\_name1\}} procedure \texttt{[time\_filter\_global1]}.
    & \\
Group 
    & Count the number of patients who received a \texttt{\{procedure\_name\}} procedure \texttt{[time\_filter\_within]} after having been diagnosed with \texttt{\{diagnosis\_name1\}} \texttt{[time\_filter\_global1]}.
    & \\
Group 
    & Count the number of \texttt{\{procedure\_name\}} procedure cases \texttt{[time\_filter\_global1]}.
    & \\
Group 
    & Count the number of patients who were prescribed \texttt{\{drug\_name\}} \texttt{[time\_filter\_global1]}.
    & \\
Group 
    & Count the number of \texttt{\{drug\_name\}} prescription cases \texttt{[time\_filter\_global1]}.
    & \\
Group 
    & Count the number of patients who were prescribed \texttt{\{drug\_name\}} \texttt{[time\_filter\_within]} after having received a \texttt{\{procedure\_name\}} procedure \texttt{[time\_filter\_global1]}.
    & \\
Group 
    & Count the number of patients who were prescribed \texttt{\{drug\_name\}} \texttt{[time\_filter\_within]} after having been diagnosed with  \texttt{\{diagnosis\_name\}} \texttt{[time\_filter\_global1]}.
    & \\
Group 
    & Count the number of patients who received a \texttt{\{lab\_name\}} lab test \texttt{[time\_filter\_global1]}.
    & \\
Group 
    & Count the number of patients who received a \texttt{\{culture\_name\}} microbiology test \texttt{[time\_filter\_global1]}.
    & \\
Group 
    & Count the number of patients who had a \texttt{\{intake\_name\}} intake \texttt{[time\_filter\_global1]}.
    & \\
Group 
    & What are\_verb the top \texttt{[n\_rank]} frequent diagnoses \texttt{[time\_filter\_global1]}?
    & \\
Group
    & What are\_verb the top \texttt{[n\_rank]} frequent diagnoses of patients aged \texttt{[age\_group]} \texttt{[time\_filter\_global1]}?
    & \\
Group 
    & What are\_verb the top \texttt{[n\_rank]} frequent diagnoses that patients were diagnosed \texttt{[time\_filter\_within]} after having received a \texttt{\{procedure\_name\}} procedure \texttt{[time\_filter\_global1]}?
    & \\
Group 
    & What are\_verb the top \texttt{[n\_rank]} frequent diagnoses that patients were diagnosed \texttt{[time\_filter\_within]} after having been diagnosed with \texttt{\{diagnosis\_name\}} \texttt{[time\_filter\_global1]}?
    & \\
Group 
    & What are\_verb the top \texttt{[n\_rank]} frequent procedures \texttt{[time\_filter\_global1]}?
    & \\
Group   
    & What are\_verb the top \texttt{[n\_rank]} frequent procedures of patients aged \texttt{[age\_group]} \texttt{[time\_filter\_global1]}?
    & \\
Group 
    & What are\_verb the top \texttt{[n\_rank]} frequent procedures that patients received \texttt{[time\_filter\_within]} after having received a \texttt{\{procedure\_name\}} procedure \texttt{[time\_filter\_global1]}?
    & \\
Group 
    & What are\_verb the top \texttt{[n\_rank]} frequent procedures that patients received \texttt{[time\_filter\_within]} after having been diagnosed with \texttt{\{diagnosis\_name\}} \texttt{[time\_filter\_global1]}?
    & \\
Group
    & What are\_verb the top \texttt{[n\_rank]} frequently prescribed drugs \texttt{[time\_filter\_global1]}?
    & \\
Group 
    & What are\_verb the top \texttt{[n\_rank]} frequently prescribed drugs of patients aged \texttt{[age\_group]} \texttt{[time\_filter\_global1]}?
    & \\
Group 
    & What are\_verb the top \texttt{[n\_rank]}  frequent prescribed drugs for patients who were also prescribed \texttt{\{drug\_name\}} \texttt{[time\_filter\_within]} \texttt{[time\_filter\_global1]}?
    & \\
Group 
    & What are\_verb the top \texttt{[n\_rank]} frequent drugs that patients were prescribed \texttt{[time\_filter\_within]} after having been prescribed with \texttt{\{drug\_name\}} \texttt{[time\_filter\_global1]}?
    & \\
Group 
    & What are\_verb the top \texttt{[n\_rank]} frequent drugs that patients were prescribed \texttt{[time\_filter\_within]} after having received a \texttt{\{procedure\_name\}} procedure \texttt{[time\_filter\_global1]}?
    & \\
Group 
    & What are\_verb the top \texttt{[n\_rank]} frequent drugs that patients were prescribed \texttt{[time\_filter\_within]} after having been diagnosed with \texttt{\{diagnosis\_name\}} \texttt{[time\_filter\_global1]}?
    & \\
Group 
    & What are\_verb the top \texttt{[n\_rank]} frequently prescribed drugs that patients aged \texttt{[age\_group]} were prescribed  \texttt{[time\_filter\_within]} after having been diagnosed with \texttt{\{diagnosis\_name\}} \texttt{[time\_filter\_global1]}?
    & \\
Group 
    & What are\_verb the top \texttt{[n\_rank]}frequently prescribed drugs that \texttt{\{gender\}} patients aged \texttt{[age\_group]} were prescribed \texttt{[time\_filter\_within]} after having been diagnosed with \texttt{\{diagnosis\_name\}} \texttt{[time\_filter\_global1]}?
    & \\
Group
    & What are\_verb the top \texttt{[n\_rank]} frequent lab test \texttt{[time\_filter\_global1]}?
    & \\
Group & 
    What are\_verb the top \texttt{[n\_rank]} frequent lab tests of patients aged  \texttt{[age\_group]} \texttt{[time\_filter\_global1]}?
    & \\
Group 
    & What are\_verb the top \texttt{[n\_rank]} frequent lab tests that patients had \texttt{[time\_filter\_within]} after having been diagnosed with \texttt{\{diagnosis\_name\}} \texttt{[time\_filter\_global1]}?
    & \\
Group 
    & What are\_verb the top \texttt{[n\_rank]}frequent lab tests that patients had \texttt{[time\_filter\_within]} after having received a \texttt{\{procedure\_name\}} procedure \texttt{[time\_filter\_global1]}?
    & \\
Group 
    & What are\_verb the top \texttt{[n\_rank]} frequent specimens tested \texttt{[time\_filter\_global1]}?
    & \\
Group 
    & What are\_verb the top \texttt{[n\_rank]} frequent specimens that patients were tested  \texttt{[time\_filter\_within]} after having been diagnosed with \texttt{\{diagnosis\_name\}} \texttt{[time\_filter\_global1]}?
    & \\
Group 
    & What are\_verb the top \texttt{[n\_rank]} frequent specimens that patients were tested \texttt{[time\_filter\_within]} after having received a \texttt{\{procedure\_name\}} procedure \texttt{[time\_filter\_global1]}?
    & \\
Group 
    & What are\_verb the top \texttt{[n\_rank]} frequent intake events \texttt{[time\_filter\_global1]}?
    & \\
Group 
    & What are\_verb the top \texttt{[n\_rank]} frequent output events \texttt{[time\_filter\_global1]}?
    & 
\end{longtblr}

%% file: supp/time_template.tex
\begin{longtblr}
[
  caption = {Full list of NL time expressions and their corresponding SQL time patterns.},
  label = {tab:time_template_full},
]
{
  colspec = {cccccX[3,c,m]X[6,l,m]},
  colsep = 0.75pt,
  rowhead = 1,
  hlines,
  rows={font=\tiny},
  rowsep=1.0pt
} 
\textbf{Time filter type} & \textbf{Expression type} & \textbf{Unit} & \textbf{Interval type} & \textbf{Option} & \textbf{NL time expression} & \textbf{SQL time pattern} \\
\texttt{global}
    & -
    & -
    & -
    & -
    & 
    &  \\
\texttt{global}
    & relative
    & hospital
    & in
    & first
    & on the first hospital visit
    & \texttt{WHERE [hospital\_dischargetime] IS NOT NULL ORDER BY [hospital\_admittime] ASC LIMIT 1} \\
\texttt{global}
    & relative
    & hospital
    & in
    & last
    & on the last hospital visit
    & \texttt{WHERE [hospital\_dischargetime] IS NOT NULL ORDER BY} \texttt{[hospital\_admittime] DESC LIMIT 1} \\
\texttt{global}
    & relative
    & hospital
    & in
    & current
    & on the current hospital visit
    & \texttt{WHERE [hospital\_dischargetime] IS NULL} \\
\texttt{global} 
    & relative
    & ICU
    & in
    & first
    & on the first ICU visit
    &\texttt{WHERE [icu\_dischargetime] IS NOT NULL ORDER BY} \texttt{[icu\_admittime] ASC LIMIT 1} \\
\texttt{global}   
    & relative
    & ICU
    & in
    & last
    & on the last ICU visit
    & \texttt{WHERE [icu\_dischargetime] IS NOT NULL ORDER BY} \texttt{[icu\_admittime] DESC LIMIT 1} \\
\texttt{global}
    & relative
    & ICU
    & in
    & current
    & on the current ICU visit
    & \texttt{WHERE [icu\_dischargetime] IS NULL} \\
\texttt{global}
    & relative
    & year
    & in
    & last
    & last year
    & \texttt{WHERE datetime([time\_column],\textquotesingle start of year\textquotesingle) = } \texttt{datetime(current\_time,\textquotesingle start of year \textquotesingle,\textquotesingle -1 year\textquotesingle)} \\
\texttt{global}
    & relative
    & year
    & until
    & last
    & until last year
    & \texttt{WHERE datetime([time\_column],\textquotesingle start of year\textquotesingle) <= }  \texttt{datetime(current\_time,\textquotesingle start of year \textquotesingle,\textquotesingle -1 year\textquotesingle)} \\
\texttt{global}
    & relative
    & year
    & since
    & last
    & since last year
    & \texttt{WHERE datetime([time\_column],\textquotesingle start of year\textquotesingle) >= }  \texttt{datetime(current\_time,\textquotesingle start of year \textquotesingle,\textquotesingle -1 year\textquotesingle)} \\
\texttt{global}
    & relative
    & year
    & in
    & this
    & this year
    & \texttt{WHERE datetime([time\_column],\textquotesingle start of year\textquotesingle) = } \texttt{datetime(current\_time,\textquotesingle start of year \textquotesingle,\textquotesingle -0 year\textquotesingle)} \\
\texttt{global}
    & relative
    & year
    & until
    & -
    & until \texttt{\{year}\} year ago
    & \texttt{WHERE datetime([time\_column]) <= datetime(current\_time,\textquotesingle -\{year\}} \texttt{year\textquotesingle)} \\
\texttt{global}
    & relative
    & year
    & since
    & -
    & since \texttt{\{year}\} year ago
    & \texttt{WHERE datetime([time\_column]) >= datetime(current\_time,\textquotesingle -\{year\}} \texttt{year\textquotesingle)} \\
\texttt{global}  
    & relative
    & month
    & in
    & last
    & last month
    & \texttt{WHERE datetime([time\_column],\textquotesingle start of month\textquotesingle) = }  \texttt{datetime(current\_time,\textquotesingle start of month\textquotesingle,\textquotesingle-1 month\textquotesingle)} \\
\texttt{global} 
    & relative
    & month
    & until
    & last
    & until last month
    & \texttt{WHERE datetime([time\_column],\textquotesingle start of month\textquotesingle) <= }  \texttt{datetime(current\_time,\textquotesingle start of month\textquotesingle,\textquotesingle-1 month\textquotesingle)} \\
\texttt{global}
    & relative
    & month
    & since
    & last
    & since last month
    & \texttt{WHERE datetime([time\_column],\textquotesingle start of month\textquotesingle) >= }  \texttt{datetime(current\_time,\textquotesingle start of month\textquotesingle,\textquotesingle-1 month\textquotesingle)} \\
 \texttt{global}       
    & relative
    & month
    & in
    & this
    & this month
    & \texttt{WHERE datetime([time\_column],\textquotesingle start of month\textquotesingle) = }  \texttt{datetime(current\_time,\textquotesingle start of month\textquotesingle,\textquotesingle-0 month\textquotesingle)} \\
 \texttt{global}       
    & relative
    & month
    & until
    & -
    & until \texttt{\{month}\} month ago
    & \texttt{WHERE datetime([time\_column]) <= } \texttt{datetime(current\_time,\textquotesingle-\{month\} month\textquotesingle)} \\
 \texttt{global}       
    & relative
    & month
    & since
    & -
    & since \texttt{\{month}\} month ago
    & \texttt{WHERE datetime([time\_column]) >= } \texttt{datetime(current\_time,\textquotesingle-\{month\} month\textquotesingle)} \\
\texttt{global}       
     & relative
    & day
    & in
    & last
    & yesterday
    & \texttt{WHERE datetime([time\_column],\textquotesingle start of day\textquotesingle) = }  \texttt{datetime(current\_time,\textquotesingle start of day\textquotesingle,\textquotesingle-1 day\textquotesingle)} \\
\texttt{global}       
    & relative
    & day
    & in
    & last
    & until yesterday
    & \texttt{WHERE datetime([time\_column],\textquotesingle start of day\textquotesingle) <= }  \texttt{datetime(current\_time,\textquotesingle start of day\textquotesingle,\textquotesingle-1 day\textquotesingle)} \\
\texttt{global}       
    & relative
    & day
    & in
    & last
    & since yesterday
    & \texttt{WHERE datetime([time\_column],\textquotesingle start of day\textquotesingle) >= }  \texttt{datetime(current\_time,\textquotesingle start of day\textquotesingle,\textquotesingle-1 day\textquotesingle)} \\
\texttt{global}       
    & relative
    & day
    & in
    & this
    & today
    & \texttt{WHERE datetime([time\_column],\textquotesingle start of day\textquotesingle) = }  \texttt{datetime(current\_time,\textquotesingle start of day\textquotesingle,\textquotesingle-0 day\textquotesingle)} \\
\texttt{global}       
    & relative
    & day
    & until
    & -
    & until \texttt{\{day}\} day ago
    & \texttt{WHERE datetime([time\_column]) <= } \texttt{datetime(current\_time,\textquotesingle-\{day\} day\textquotesingle)} \\
 \texttt{global}       
    & relative
    & day
    & since
    & -
    & since \texttt{\{day}\} day ago
    & \texttt{WHERE datetime([time\_column]) >= } \texttt{datetime(current\_time,\textquotesingle-\{day\} day\textquotesingle)} \\
\texttt{global}       
    & absolute
    & year
    & in
    & -
    & in \texttt{\{year}\}
    & \texttt{WHERE strftime(\textquotesingle\%Y\textquotesingle,[time\_column])= } \textquotesingle\texttt{\{year\}}\textquotesingle \\
\texttt{global}       
    & absolute
    & year
    & until
    & -
    & until \texttt{\{year}\} 
    & \texttt{WHERE strftime(\textquotesingle\%Y\textquotesingle,[time\_column]) <= } \textquotesingle\texttt{\{year\}}\textquotesingle \\
\texttt{global}       
    & absolute
    & year
    & since
    & -
    & since \texttt{\{year}\}
    & \texttt{WHERE strftime(\textquotesingle\%Y\textquotesingle,[time\_column]) >= } \textquotesingle\texttt{\{year\}}\textquotesingle \\
\texttt{global}       
    & absolute
    & month
    & in
    & -
    & in \texttt{\{month}\}/\texttt{\{year}\}
    & \texttt{WHERE strftime(\textquotesingle\%Y-\%m\textquotesingle,[time\_column])= } \textquotesingle\texttt{\{year\}}-\texttt{\{month\}}\textquotesingle \\
\texttt{global}       
    & absolute
    & month
    & until
    & -
    & until \texttt{\{month}\}/\texttt{\{year}\}
    & \texttt{WHERE strftime(\textquotesingle\%Y-\%m\textquotesingle,[time\_column]) <= } \textquotesingle\texttt{\{year\}}-\texttt{\{month\}}\textquotesingle \\
\texttt{global}       
    & absolute
    & month
    & since
    & -
    & since \texttt{\{month}\}/\texttt{\{year}\}
    & \texttt{WHERE strftime(\textquotesingle\%Y-\%m\textquotesingle,[time\_column]) >= } \textquotesingle\texttt{\{year\}}-\texttt{\{month\}}\textquotesingle \\ 
\texttt{global}       
    & absolute
    & day
    & in
    & -
    & on \texttt{\{month}\}/\texttt{\{day}\}/\texttt{\{year}\}
    & \texttt{WHERE strftime(\textquotesingle\%Y-\%m-\%d\textquotesingle,[time\_column]) = } \textquotesingle\texttt{\{year\}}-\texttt{\{month\}}-\texttt{\{day\}}\textquotesingle \\
\texttt{global}       
    & absolute
    & day
    & until
    & -
    & until \texttt{\{month}\}/\texttt{\{day}\}/\texttt{\{year}\}
    & \texttt{WHERE strftime(\textquotesingle\%Y-\%m-\%d\textquotesingle,[time\_column]) <= } \textquotesingle\texttt{\{year\}}-\texttt{\{month\}}-\texttt{\{day\}}\textquotesingle \\
\texttt{global}       
    & absolute
    & day
    & since
    & -
    & since \texttt{\{month}\}/\texttt{\{day}\}/\texttt{\{year}\}
    & \texttt{WHERE strftime(\textquotesingle\%Y-\%m-\%d\textquotesingle,[time\_column]) >= } \textquotesingle\texttt{\{year\}}-\texttt{\{month\}}\texttt{\{day\}}\textquotesingle \\ 
\texttt{global}       
    & mix
    & month
    & in
    & last
    & in \texttt{\{month}\}/last year
    & \texttt{WHERE datetime([time\_column],\textquotesingle start of year\textquotesingle) = } \texttt{datetime(current\_time,\textquotesingle start of year\textquotesingle,\textquotesingle-1 year\textquotesingle) AND} \texttt{ strftime(\textquotesingle\%m\textquotesingle,[time\_column]) = \textquotesingle\texttt{\{month\}}\textquotesingle}\\
\texttt{global}       
    & mix
    & month
    & in
    & this
    & in \texttt{\{month}\}/this year
    & \texttt{WHERE datetime([time\_column],\textquotesingle start of year\textquotesingle) = } \texttt{datetime(current\_time,\textquotesingle start of year\textquotesingle,\textquotesingle-0 year\textquotesingle) AND} \texttt{ strftime(\textquotesingle\%m\textquotesingle,[time\_column]) = \textquotesingle\texttt{\{month\}}\textquotesingle}\\
\texttt{global}       
    & mix
    & day
    & in
    & last
    & on \texttt{\{month}\}/\texttt{\{day}\}/last year
    & \texttt{WHERE datetime([time\_column],\textquotesingle start of year\textquotesingle) = } \texttt{datetime(current\_time,\textquotesingle start of year\textquotesingle,\textquotesingle-1 year\textquotesingle) AND} \texttt{ strftime(\textquotesingle\%m-\%d\textquotesingle,[time\_column]) = \textquotesingle\texttt{\{month\}}-\texttt{\{day\}}\textquotesingle}\\
\texttt{global}       
    & mix
    & day
    & in
    & this
    & on \texttt{\{month}\}/\texttt{\{day}\}/this year
    & \texttt{WHERE datetime([time\_column],\textquotesingle start of year\textquotesingle) = } \texttt{datetime(current\_time,\textquotesingle start of year\textquotesingle,\textquotesingle-0 year\textquotesingle) AND} \texttt{ strftime(\textquotesingle\%m-\%d\textquotesingle,[time\_column]) = \textquotesingle\texttt{\{month\}}-\texttt{\{day\}}\textquotesingle}\\
\texttt{global}       
    & mix
    & day
    & in
    & last
    & on last month/\texttt{\{day}\}
    & \texttt{WHERE datetime([time\_column],\textquotesingle start of month\textquotesingle) = } \texttt{datetime(current\_time,\textquotesingle start of month\textquotesingle,\textquotesingle-1 month\textquotesingle) AND} \texttt{ strftime(\textquotesingle\%d\textquotesingle,[time\_column]) = \textquotesingle\texttt{\{day\}}\textquotesingle}\\
\texttt{global}       
    & mix
    & day
    & in
    & this
    & on this month/\texttt{\{day}\}
    & \texttt{WHERE datetime([time\_column],\textquotesingle start of month\textquotesingle) = } \texttt{datetime(current\_time,\textquotesingle start of month\textquotesingle,\textquotesingle-0 month\textquotesingle) AND} \texttt{ strftime(\textquotesingle\%d\textquotesingle,[time\_column]) = \textquotesingle\texttt{\{day\}}\textquotesingle}\\
\texttt{within}       
    & -
    & -
    & -
    & -
    & 
    &  \\
\texttt{within}       
    & -
    & hospital
    & in
    & -
    & within the same hospital visit
    & \texttt{WHERE [hospital\_admission\_id1] = [hospital\_admission\_id2] }\\
\texttt{within}       
    & -
    & ICU
    & in
    & -
    & within the same icu visit
    & \texttt{WHERE [icu\_admission\_id1] = [icu\_admission\_id2] }\\
\texttt{within}       
    & -
    & year
    & in
    & -
    & within the same year
    &  \texttt{WHERE datetime([time\_column1],\textquotesingle start of year\textquotesingle) = } \texttt{datetime([time\_column2],\textquotesingle start of year\textquotesingle}\\ 
\texttt{within}       
    & -
    & n\_year
    & in
    & -
    & within \texttt{\{year}\} year
    &  \texttt{WHERE datetime([time\_column2]) BETWEEN } \texttt{datetime([time\_column1] AND} \texttt{datetime([time\_column1],\textquotesingle +\texttt{\{year}\} year\textquotesingle}\\ 
\texttt{within}       
    & -
    & month
    & in
    & -
    & within the same month
    &  \texttt{WHERE datetime([time\_column1],\textquotesingle start of month\textquotesingle) = } \texttt{datetime([time\_column2],\textquotesingle start of month\textquotesingle}\\ 
\texttt{within}       
    & -
    & n\_month
    & in
    & -
    & within \texttt{\{month}\} month
    &  \texttt{WHERE datetime([time\_column2]) BETWEEN } \texttt{datetime([time\_column1] AND} \texttt{datetime([time\_column1],\textquotesingle +\texttt{\{month}\} month\textquotesingle}\\ 
\texttt{within}       
    & -
    & day
    & in
    & -
    & within the same day
    &  \texttt{WHERE datetime([time\_column1],\textquotesingle start of day\textquotesingle) = } \texttt{datetime([time\_column2],\textquotesingle start of day\textquotesingle}\\ 
\texttt{within}       
    & -
    & n\_day
    & in
    & -
    & within \texttt{\{day}\} day
    &  \texttt{WHERE datetime([time\_column2]) BETWEEN } \texttt{datetime([time\_column1] AND} \texttt{datetime([time\_column1],\textquotesingle +\texttt{\{day}\} day\textquotesingle}\\ 
\texttt{within}       
    & -
    & exact
    & in
    & -
    & at the same time
    &  \texttt{WHERE datetime([time\_column1]) = datetime([time\_column2])}  \\
\texttt{exact}       
    & relative
    & exact
    & at
    & -
    & first
    &  \texttt{ORDER BY [time\_column] ASC LIMIT 1 }  \\
\texttt{exact}       
    & relative
    & exact
    & at
    & -
    & second
    &  \texttt{ORDER BY [time\_column] ASC LIMIT 1 OFFSET 1}  \\
\texttt{exact}       
    & relative
    & exact
    & at
    & -
    & second to last
    &  \texttt{ORDER BY [time\_column] DESC LIMIT 1 OFFSET 1}  \\
\texttt{exact}       
    & relative
    & exact
    & at
    & -
    & last
    &  \texttt{ORDER BY [time\_column1] DESC LIMIT 1 }  \\
\texttt{exact}       
    & absolute
    & exact
    & at
    & -
    &
    at \texttt{\{year}\}-\texttt{\{month}\}-\texttt{\{day}\} 
    \texttt{\{hour}\} :\texttt{\{minute}\}:\texttt{\{second}\}
    &  \texttt{WHERE datetime([time\_column]) = \textquotesingle\texttt{\{year}\}-\texttt{\{month}\}-\texttt{\{day}\} \texttt{\{hour}\}:\texttt{\{minute}\} :\texttt{\{second}\}\textquotesingle}  \\
\end{longtblr}

%% file: supp/schema_mapping.tex
\begin{table}[h!]
\centering
\caption{Schema mapping in both MIMIC-III and eICU.}
\scalebox{0.7}{
\begin{tabular}{ccc}
\toprule
\textbf{Condition value slots} 
    & \textbf{MIMIC-III} 
    & \textbf{eICU} \\
\midrule
\multirow{2}{*}{\texttt{\{abbreviation\}}} 
    & d\_icd\_procedures.short\_title & \multirow{2}{*}{-} \\
    & d\_icd\_diagnoses.short\_title &  \\
\midrule    
\texttt{\{admission\_route\}} 
    & admissions.admission\_location
    & patient.hospitaladmitsource \\
\midrule        
\texttt{\{careunit\}} 
    & transfers.curr\_careunit
    & - \\
\midrule        
\texttt{\{culture\_name\}} 
    & microbiologyevents.spec\_type\_desc
    & microlab.culturesite \\
\midrule        
\texttt{\{diagnosis\_name\}} 
    & d\_icd\_diagnoses.short\_title
    & diagnosis.diagnosisname \\
\midrule        
\texttt{\{drug\_name\}} 
    & prescriptions.drug\_name
    & medication.drugname \\
\midrule        
\texttt{\{drug\_route\}} 
    & prescriptions.route
    & medication.routeadmin \\    
\midrule        
\texttt{\{gender\}} 
    & patients.gender
    & patient.gender \\    
\midrule        
\texttt{\{intake\_name\}} 
    & d\_items.label
    & intakeoutput.celllabel \\ 
\midrule        
\texttt{\{lab\_name\}} 
    & d\_labitems.label
    & lab.labname \\
\midrule        
\texttt{\{lab\_value\}} 
    & labevents.valuenum
    & lab.labresult \\        
\midrule        
\texttt{\{output\_name\}} 
    & d\_items.label
    & intakeoutput.celllabel \\    
\midrule        
\texttt{\{patient\_id\}} 
    & patients.subject\_id
    & patient.uniquepid \\         
\midrule        
\texttt{\{procedure\_name\}} 
    & d\_icd\_procedures.short\_title
    & treatment.treatmentname \\     
\midrule        
\texttt{\{vital\_name\}} 
    & d\_items.label
    & - \\
\midrule        
\multirow{7}{*}{\texttt{\{vital\_value\}}} 
    & \multirow{7}{*}{chartevents.valuenum} 
    & vitalperiodic.temperature,  \\
    & & vitalperiodic.sao2,  \\
    & & vitalperiodic.heartrate,  \\
    & & vitalperiodic.respiration, \\
    & & vitalperiodic.systemicsystolic,  \\   
    & & vitalperiodic.systemicdiastolic, \\   
    & & vitalperiodic.systemicmean \\
\midrule        
\texttt{\{ward\_id\}} 
    & transfers.curr\_wardid
    & patient.wardid \\
\bottomrule
\end{tabular}
}
\label{tab:schema_mapping}
\end{table}

%% file: references.bib
@article{johnson2016mimic,
  title={MIMIC-III, a freely accessible critical care database},
  author={Johnson, Alistair EW and Pollard, Tom J and Shen, Lu and Li-Wei, H Lehman and Feng, Mengling and Ghassemi, Mohammad and Moody, Benjamin and Szolovits, Peter and Celi, Leo Anthony and Mark, Roger G},
  journal={Scientific data},
  volume={3},
  number={1},
  pages={1--9},
  year={2016},
  publisher={Nature Publishing Group}
}

@article{pollard2018eicu,
  title={The eICU Collaborative Research Database, a freely available multi-center database for critical care research},
  author={Pollard, Tom J and Johnson, Alistair EW and Raffa, Jesse D and Celi, Leo A and Mark, Roger G and Badawi, Omar},
  journal={Scientific data},
  volume={5},
  number={1},
  pages={1--13},
  year={2018},
  publisher={Nature Publishing Group}
}

@article{johnson2019mimic,
  title={MIMIC-CXR, a de-identified publicly available database of chest radiographs with free-text reports},
  author={Johnson, Alistair EW and Pollard, Tom J and Berkowitz, Seth J and Greenbaum, Nathaniel R and Lungren, Matthew P and Deng, Chih-ying and Mark, Roger G and Horng, Steven},
  journal={Scientific data},
  volume={6},
  number={1},
  pages={1--8},
  year={2019},
  publisher={Nature Publishing Group}
}

@article{johnson2021mimic,
  title={MIMIC-IV (version 1.0)},
  author={Johnson, Alistair and Bulgarelli, Lucas and Pollard, Tom and Horng, Steven and Celi, Leo Anthony and Mark, Roger},
  journal={PhysioNet},
  year={2021},
  doi={https://doi.org/10.13026/s6n6-xd98}
}

@article{goldberger2000physiobank,
  title={PhysioBank, PhysioToolkit, and PhysioNet: components of a new research resource for complex physiologic signals},
  author={Goldberger, Ary L and Amaral, Luis AN and Glass, Leon and Hausdorff, Jeffrey M and Ivanov, Plamen Ch and Mark, Roger G and Mietus, Joseph E and Moody, George B and Peng, Chung-Kang and Stanley, H Eugene},
  journal={circulation},
  volume={101},
  number={23},
  pages={e215--e220},
  year={2000},
  publisher={Am Heart Assoc}
}

@article{liu2019roberta,
  title={Roberta: A robustly optimized bert pretraining approach},
  author={Liu, Yinhan and Ott, Myle and Goyal, Naman and Du, Jingfei and Joshi, Mandar and Chen, Danqi and Levy, Omer and Lewis, Mike and Zettlemoyer, Luke and Stoyanov, Veselin},
  journal={arXiv preprint arXiv:1907.11692},
  year={2019}
}

@article{gao2020pile,
  title={The Pile: An 800GB Dataset of Diverse Text for Language Modeling},
  author={Gao, Leo and Biderman, Stella and Black, Sid and Golding, Laurence and Hoppe, Travis and Foster, Charles and Phang, Jason and He, Horace and Thite, Anish and Nabeshima, Noa and others},
  journal={arXiv preprint arXiv:2101.00027},
  year={2020}
}

@article{zhong2017seq2sql,
  title={Seq2sql: Generating structured queries from natural language using reinforcement learning},
  author={Zhong, Victor and Xiong, Caiming and Socher, Richard},
  journal={arXiv preprint arXiv:1709.00103},
  year={2017}
}

@inproceedings{yu2018spider,
  title={Spider: A Large-Scale Human-Labeled Dataset for Complex and Cross-Domain Semantic Parsing and Text-to-SQL Task},
  author={Yu, Tao and Zhang, Rui and Yang, Kai and Yasunaga, Michihiro and Wang, Dongxu and Li, Zifan and Ma, James and Li, Irene and Yao, Qingning and Roman, Shanelle and others},
  booktitle={Proceedings of the 2018 Conference on Empirical Methods in Natural Language Processing},
  pages={3911--3921},
  year={2018}
}

@article{lee2021kaggledbqa,
  title={KaggleDBQA: Realistic Evaluation of Text-to-SQL Parsers},
  author={Lee, Chia-Hsuan and Polozov, Oleksandr and Richardson, Matthew},
  journal={arXiv preprint arXiv:2106.11455},
  year={2021}
}

@article{hazoom2021text,
  title={Text-to-SQL in the Wild: A Naturally-Occurring Dataset Based on Stack Exchange Data},
  author={Hazoom, Moshe and Malik, Vibhor and Bogin, Ben},
  journal={arXiv preprint arXiv:2106.05006},
  year={2021}
}

@inproceedings{shi2021learning,
  title={Learning Contextual Representations for Semantic Parsing with Generation-Augmented Pre-Training},
  author={Shi, Peng and Ng, Patrick and Wang, Zhiguo and Zhu, Henghui and Li, Alexander Hanbo and Wang, Jun and dos Santos, Cicero Nogueira and Xiang, Bing},
  booktitle={Proceedings of the AAAI Conference on Artificial Intelligence},
  volume={35},
  number={15},
  pages={13806--13814},
  year={2021}
}

@inproceedings{pampari2018emrqa,
  title={emrQA: A Large Corpus for Question Answering on Electronic Medical Records},
  author={Pampari, Anusri and Raghavan, Preethi and Liang, Jennifer and Peng, Jian},
  booktitle={Proceedings of the 2018 Conference on Empirical Methods in Natural Language Processing},
  pages={2357--2368},
  year={2018}
}

@inproceedings{wang2020text,
  title={Text-to-sql generation for question answering on electronic medical records},
  author={Wang, Ping and Shi, Tian and Reddy, Chandan K},
  booktitle={Proceedings of The Web Conference 2020},
  pages={350--361},
  year={2020}
}

@inproceedings{raghavan2021emrkbqa,
  title={emrKBQA: A Clinical Knowledge-Base Question Answering Dataset},
  author={Raghavan, Preethi and Liang, Jennifer J and Mahajan, Diwakar and Chandra, Rachita and Szolovits, Peter},
  booktitle={Proceedings of the 20th Workshop on Biomedical Language Processing},
  pages={64--73},
  year={2021}
}

@inproceedings{park2021knowledge,
  title={Knowledge graph-based question answering with electronic health records},
  author={Park, Junwoo and Cho, Youngwoo and Lee, Haneol and Choo, Jaegul and Choi, Edward},
  booktitle={Machine Learning for Healthcare Conference},
  pages={36--53},
  year={2021},
  organization={PMLR}
}

@article{bardhan2022drugehrqa,
  title={DrugEHRQA: A Question Answering Dataset on Structured and Unstructured Electronic Health Records For Medicine Related Queries},
  author={Bardhan, Jayetri and Colas, Anthony and Roberts, Kirk and Wang, Daisy Zhe},
  journal={arXiv preprint arXiv:2205.01290},
  year={2022}
}

@inproceedings{hemphill-etal-1990-atis,
    title = "The {ATIS} Spoken Language Systems Pilot Corpus",
    author = "Hemphill, Charles T.  and
      Godfrey, John J.  and
      Doddington, George R.",
    booktitle = "Speech and Natural Language: Proceedings of a Workshop Held at Hidden Valley, {P}ennsylvania, June 24-27,1990",
    year = "1990",
    url = "https://aclanthology.org/H90-1021",
}

@inproceedings{zelle1996learning,
  title={Learning to parse database queries using inductive logic programming},
  author={Zelle, John M and Mooney, Raymond J},
  booktitle={Proceedings of the national conference on artificial intelligence},
  pages={1050--1055},
  year={1996}
}

@inproceedings{suhr2020exploring,
  title={Exploring unexplored generalization challenges for cross-database semantic parsing},
  author={Suhr, Alane and Chang, Ming-Wei and Shaw, Peter and Lee, Kenton},
  booktitle={Proceedings of the 58th Annual Meeting of the Association for Computational Linguistics},
  pages={8372--8388},
  year={2020}
}

@inproceedings{yu2019sparc,
  title={SParC: Cross-Domain Semantic Parsing in Context},
  author={Yu, Tao and Zhang, Rui and Yasunaga, Michihiro and Tan, Yi Chern and Lin, Xi Victoria and Li, Suyi and Er, Heyang and Li, Irene and Pang, Bo and Chen, Tao and others},
  booktitle={Proceedings of the 57th Annual Meeting of the Association for Computational Linguistics},
  pages={4511--4523},
  year={2019}
}

@article{yaghmazadeh2017sqlizer,
  title={SQLizer: query synthesis from natural language},
  author={Yaghmazadeh, Navid and Wang, Yuepeng and Dillig, Isil and Dillig, Thomas},
  journal={Proceedings of the ACM on Programming Languages},
  volume={1},
  number={OOPSLA},
  pages={1--26},
  year={2017},
  publisher={ACM New York, NY, USA}
}

@article{finegan2018improving,
  title={Improving text-to-sql evaluation methodology},
  author={Finegan-Dollak, Catherine and Kummerfeld, Jonathan K and Zhang, Li and Ramanathan, Karthik and Sadasivam, Sesh and Zhang, Rui and Radev, Dragomir},
  journal={arXiv preprint arXiv:1806.09029},
  year={2018}
}

@article{zhang2020did,
  title={Did you ask a good question? a cross-domain question intention classification benchmark for text-to-sql},
  author={Zhang, Yusen and Dong, Xiangyu and Chang, Shuaichen and Yu, Tao and Shi, Peng and Zhang, Rui},
  journal={arXiv preprint arXiv:2010.12634},
  year={2020}
}

@article{rajpurkar2018know,
  title={Know what you don't know: Unanswerable questions for SQuAD},
  author={Rajpurkar, Pranav and Jia, Robin and Liang, Percy},
  journal={arXiv preprint arXiv:1806.03822},
  year={2018}
}

@inproceedings{clark2017simple,
    title = "Simple and Effective Multi-Paragraph Reading Comprehension",
    author = "Clark, Christopher  and
      Gardner, Matt",
    booktitle = "Proceedings of the 56th Annual Meeting of the Association for Computational Linguistics (Volume 1: Long Papers)",
    month = jul,
    year = "2018",
    address = "Melbourne, Australia",
    publisher = "Association for Computational Linguistics",
    url = "https://aclanthology.org/P18-1078",
    doi = "10.18653/v1/P18-1078",
    pages = "845--855",
}

@inproceedings{joshi-etal-2017-triviaqa,
    title = "{T}rivia{QA}: A Large Scale Distantly Supervised Challenge Dataset for Reading Comprehension",
    author = "Joshi, Mandar  and
      Choi, Eunsol  and
      Weld, Daniel  and
      Zettlemoyer, Luke",
    booktitle = "Proceedings of the 55th Annual Meeting of the Association for Computational Linguistics (Volume 1: Long Papers)",
    month = jul,
    year = "2017",
    address = "Vancouver, Canada",
    publisher = "Association for Computational Linguistics",
    url = "https://aclanthology.org/P17-1147",
    doi = "10.18653/v1/P17-1147",
    pages = "1601--1611",
}

@inproceedings{jia-liang-2017-adversarial,
    title = "Adversarial Examples for Evaluating Reading Comprehension Systems",
    author = "Jia, Robin  and
      Liang, Percy",
    booktitle = "Proceedings of the 2017 Conference on Empirical Methods in Natural Language Processing",
    month = sep,
    year = "2017",
    address = "Copenhagen, Denmark",
    publisher = "Association for Computational Linguistics",
    url = "https://aclanthology.org/D17-1215",
    doi = "10.18653/v1/D17-1215",
    pages = "2021--2031",
}

@misc{prithivida2021styleformer,
  author = {Prithiviraj Damodaran},
  title = {Styleformer},
  year = {2021},
  publisher = {GitHub},
  journal = {GitHub repository},
  url = "https://github.com/PrithivirajDamodaran/Styleformer"
}

@misc{prithivida2021parrot,
  author       = {Prithiviraj Damodaran},
  title        = {Parrot: Paraphrase generation for NLU},
  year         = {2021},
  version      = {v1.0},
  journal = {GitHub repository},  
  url = "https://github.com/PrithivirajDamodaran/Parrot_Paraphraser"  
}

@misc{alisetti2021t5,
  author       = {Sai Vamsi Alisetti},
  title        = {Paraphrase Generator with T5},
  year         = {2020},
  journal = {GitHub repository},  
  url = "https://github.com/Vamsi995/Paraphrase-Generator"  
}

@article{fan2021beyond,
  title={Beyond english-centric multilingual machine translation},
  author={Fan, Angela and Bhosale, Shruti and Schwenk, Holger and Ma, Zhiyi and El-Kishky, Ahmed and Goyal, Siddharth and Baines, Mandeep and Celebi, Onur and Wenzek, Guillaume and Chaudhary, Vishrav and others},
  journal={Journal of Machine Learning Research},
  volume={22},
  number={107},
  pages={1--48},
  year={2021}
}

@article{tang2020multilingual,
  title={Multilingual translation with extensible multilingual pretraining and finetuning},
  author={Tang, Yuqing and Tran, Chau and Li, Xian and Chen, Peng-Jen and Goyal, Naman and Chaudhary, Vishrav and Gu, Jiatao and Fan, Angela},
  journal={arXiv preprint arXiv:2008.00401},
  year={2020}
}

@article{zheng2020out,
  title={Out-of-domain detection for natural language understanding in dialog systems},
  author={Zheng, Yinhe and Chen, Guanyi and Huang, Minlie},
  journal={IEEE/ACM Transactions on Audio, Speech, and Language Processing},
  volume={28},
  pages={1198--1209},
  year={2020},
  publisher={IEEE}
}

@article{marek2021oodgan,
  title={Oodgan: Generative adversarial network for out-of-domain data generation},
  author={Marek, Petr and Naik, Vishal Ishwar and Auvray, Vincent and Goyal, Anuj},
  journal={arXiv preprint arXiv:2104.02484},
  year={2021}
}

@article{raffel2019exploring,
  title={Exploring the limits of transfer learning with a unified text-to-text transformer},
  author={Raffel, Colin and Shazeer, Noam and Roberts, Adam and Lee, Katherine and Narang, Sharan and Matena, Michael and Zhou, Yanqi and Li, Wei and Liu, Peter J},
  journal={arXiv preprint arXiv:1910.10683},
  year={2019}
}

@article{malinin2020uncertainty,
  title={Uncertainty estimation in autoregressive structured prediction},
  author={Malinin, Andrey and Gales, Mark},
  journal={arXiv preprint arXiv:2002.07650},
  year={2020}
}

@inproceedings{xiao2021hallucination,
  title={On Hallucination and Predictive Uncertainty in Conditional Language Generation},
  author={Xiao, Yijun and Wang, William Yang},
  booktitle={Proceedings of the 16th Conference of the European Chapter of the Association for Computational Linguistics: Main Volume},
  pages={2734--2744},
  year={2021}
}

@InProceedings{pmlr-v158-bae21a,
  title = 	 {Question Answering for Complex Electronic Health Records Database using Unified Encoder-Decoder Architecture},
  author =       {Bae, Seongsu and Kim, Daeyoung and Kim, Jiho and Choi, Edward},
  booktitle = 	 {Proceedings of Machine Learning for Health},
  pages = 	 {13--25},
  year = 	 {2021},
  volume = 	 {158},
  series = 	 {Proceedings of Machine Learning Research},
  month = 	 {04 Dec},
  publisher =    {PMLR},
  url = 	 {https://proceedings.mlr.press/v158/bae21a.html},
}

@inproceedings{elgohary-etal-2020-speak,
    title = "Speak to your Parser: Interactive Text-to-{SQL} with Natural Language Feedback",
    author = "Elgohary, Ahmed  and
      Hosseini, Saghar  and
      Hassan Awadallah, Ahmed",
    booktitle = "Proceedings of the 58th Annual Meeting of the Association for Computational Linguistics",
    month = jul,
    year = "2020",
    address = "Online",
    publisher = "Association for Computational Linguistics",
    url = "https://aclanthology.org/2020.acl-main.187",
    doi = "10.18653/v1/2020.acl-main.187",
    pages = "2065--2077",
}
